\RequirePackage{fix-cm}
\documentclass[twocolumn]{svjour3} 
\smartqed  
\usepackage{graphicx}
\usepackage[numbers]{natbib}
\usepackage{framed,multirow}
\usepackage{amssymb}
\usepackage{latexsym}
\usepackage{url}
\usepackage{xcolor}
\usepackage{ulem}
\usepackage{array}
\usepackage{comment}
\definecolor{newcolor}{rgb}{.8,.349,.1}

\newcolumntype{L}[1]{>{\raggedright\let\newline\\\arraybackslash\hspace{0pt}}m{#1}}
\newcolumntype{C}[1]{>{\centering\let\newline\\\arraybackslash\hspace{0pt}}m{#1}}
\newcolumntype{R}[1]{>{\raggedleft\let\newline\\\arraybackslash\hspace{0pt}}m{#1}}
\usepackage{hhline}
\usepackage{amsmath}
\usepackage[super]{nth}

\begin{document}

\sloppy

\title{MLMT-CNN for Object Detection and Segmentation in Multi-layer and Multi-spectral Images}

\author{Majedaldein {Almahasneh} \and
        Adeline {Paiement} \and
        Xianghua {Xie} \and
        Jean {Aboudarham} \and
}

\institute{M. {Almahasneh} \at
              Department of Computer Science, Swansea University, Swansea, UK \\
              \email{809508@swansea.ac.uk}
           \and
              A. {Paiement} \at
              Université de Toulon, Aix Marseille Univ, CNRS, LIS, Marseille, France \\
              \email{adeline.paiement@univ-tln.fr}
           \and
              X. {Xie} \at
              Department of Computer Science, Swansea University, Swansea, UK \\
              \email{x.xie@swansea.ac.uk}
           \and
              J. {Aboudarham} \at
              Observatoire de Paris/PSL, Paris, France \\
              \email{Jean.Aboudarham@obspm.fr}
}


\maketitle

\begin{abstract}
Precisely localising solar Active Regions (AR) from multi-spectral images is a challenging but important task in understanding solar activity and its influence on space weather. A main challenge comes from each modality capturing a different location of the 3D objects, as opposed to typical multi-spectral imaging scenarios where all image bands observe the same scene.  {Thus, we refer to this special multi-spectral scenario as \textit{multi-layer}.}
We present a multi-task deep learning framework that exploits the dependencies between image bands to produce 3D AR localisation (segmentation and detection) where different image bands (and physical locations) have their own set of results.
Furthermore, to address the difficulty of producing dense AR annotations for training supervised machine learning (ML) algorithms, we adapt a training strategy based on weak labels (i.e. bounding boxes) in a recursive manner. 
We compare our detection and segmentation stages against baseline approaches for solar image analysis (multi-channel coronal hole detection, SPOCA for ARs)
and state-of-the-art deep learning methods (Faster RCNN, U-Net).
Additionally, both detection and segmentation stages are quantitatively validated on artificially created data of similar spatial configurations made from annotated multi-modal magnetic resonance images.
{Our framework achieves an average of 0.72 IoU (segmentation) and 0.90 F1 score (detection) across all modalities, comparing to the best performing baseline methods with scores of 0.53 and 0.58, respectively, on the artificial dataset,
and 0.84 F1 score in the AR detection task comparing to baseline of 0.82 F1 score.}
Our segmentation results are qualitatively validated by an expert on real ARs.

\keywords{Image segmentation \and object detection \and deep learning \and weakly supervised learning \and multi-spectral images \and solar image analysis \and solar active regions}

\end{abstract}

\section{Introduction}
\label{intro}

\begin{figure}
\centering
\begin{tabular}{ccc}

\includegraphics[width=0.30\linewidth]{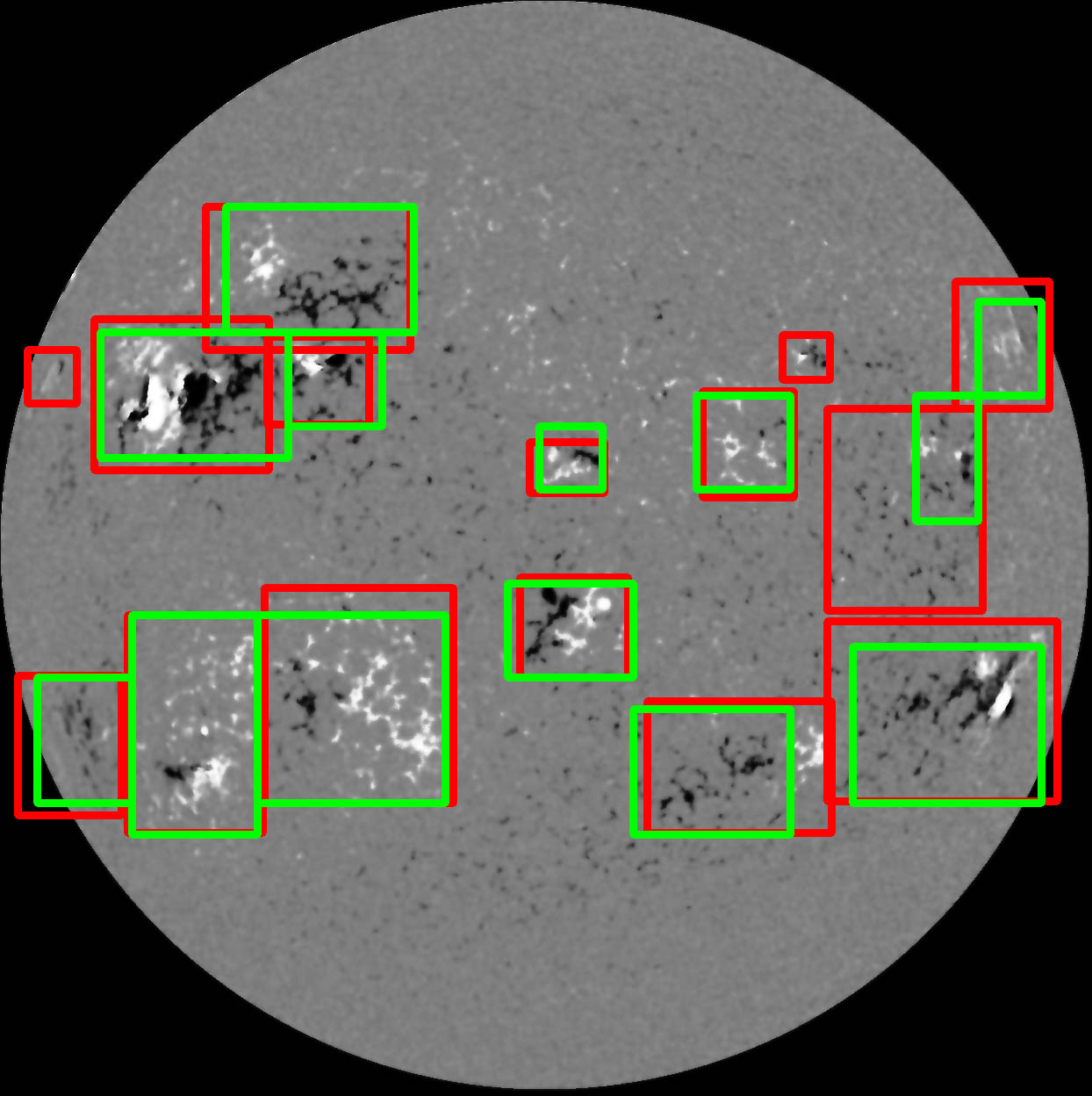} 
\includegraphics[width=0.30\linewidth]{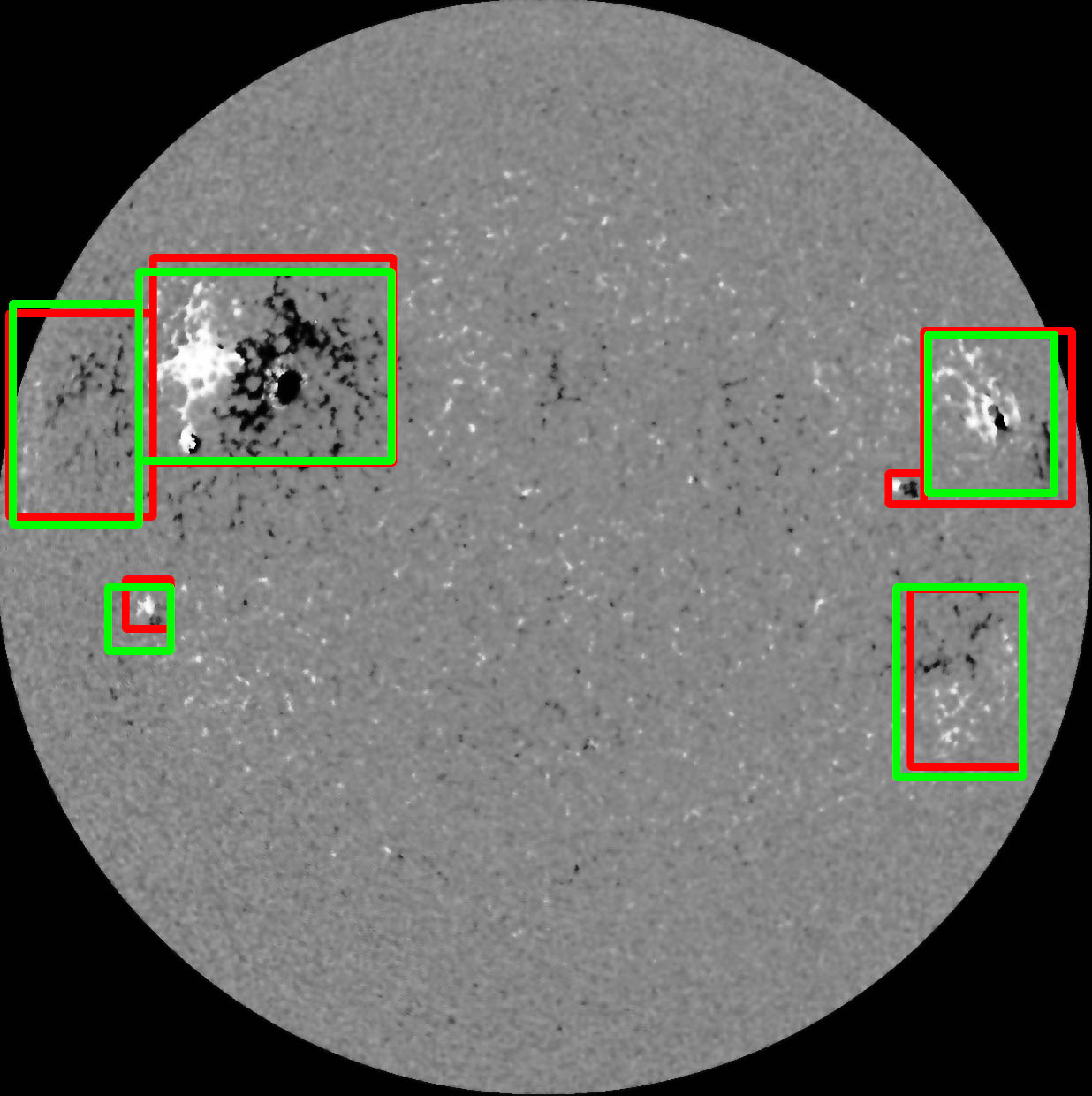} 
\includegraphics[width=0.30\linewidth]{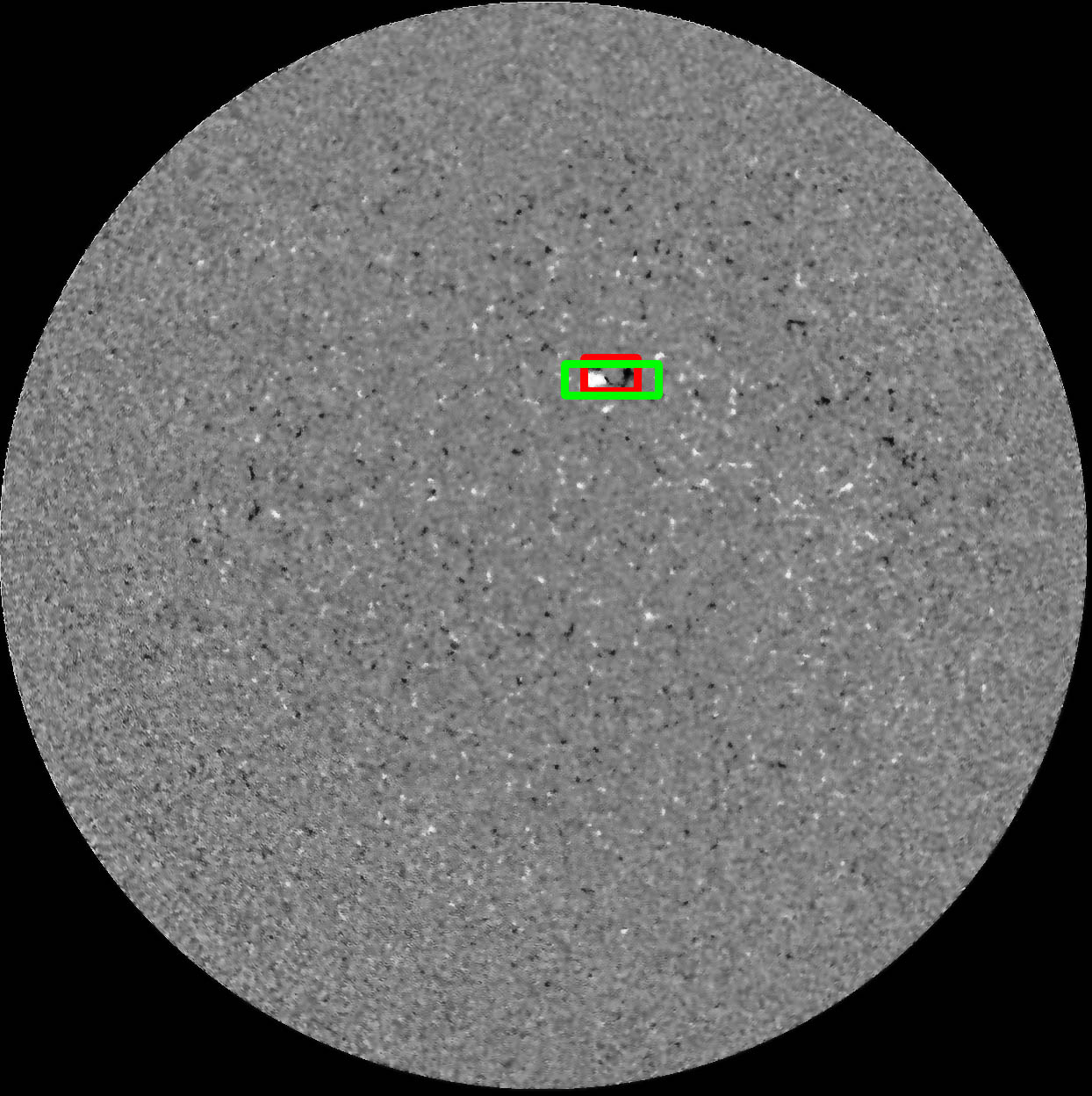} \\

\includegraphics[width=0.30\linewidth]{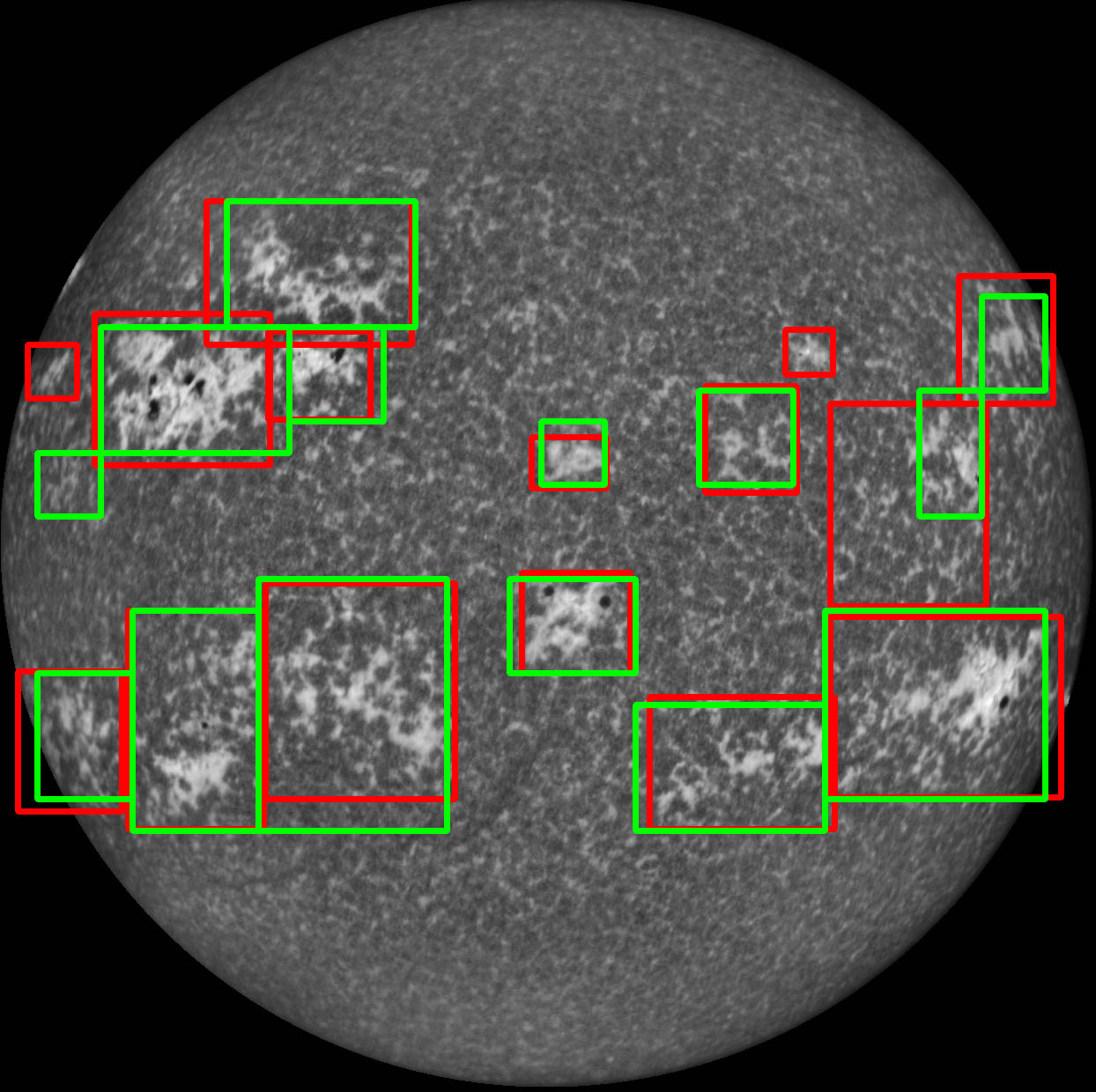} 
\includegraphics[width=0.30\linewidth]{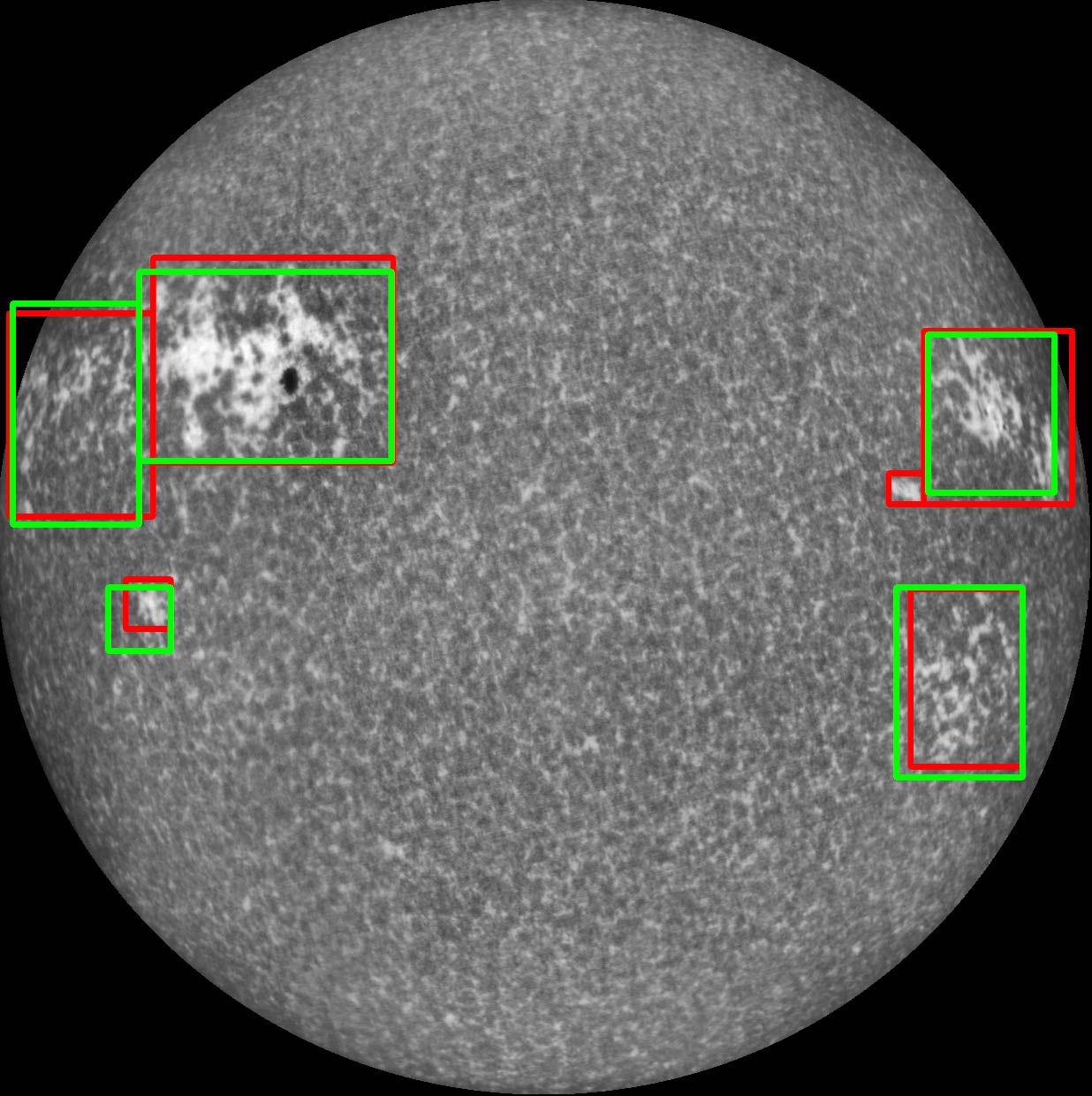} 
\includegraphics[width=0.30\linewidth]{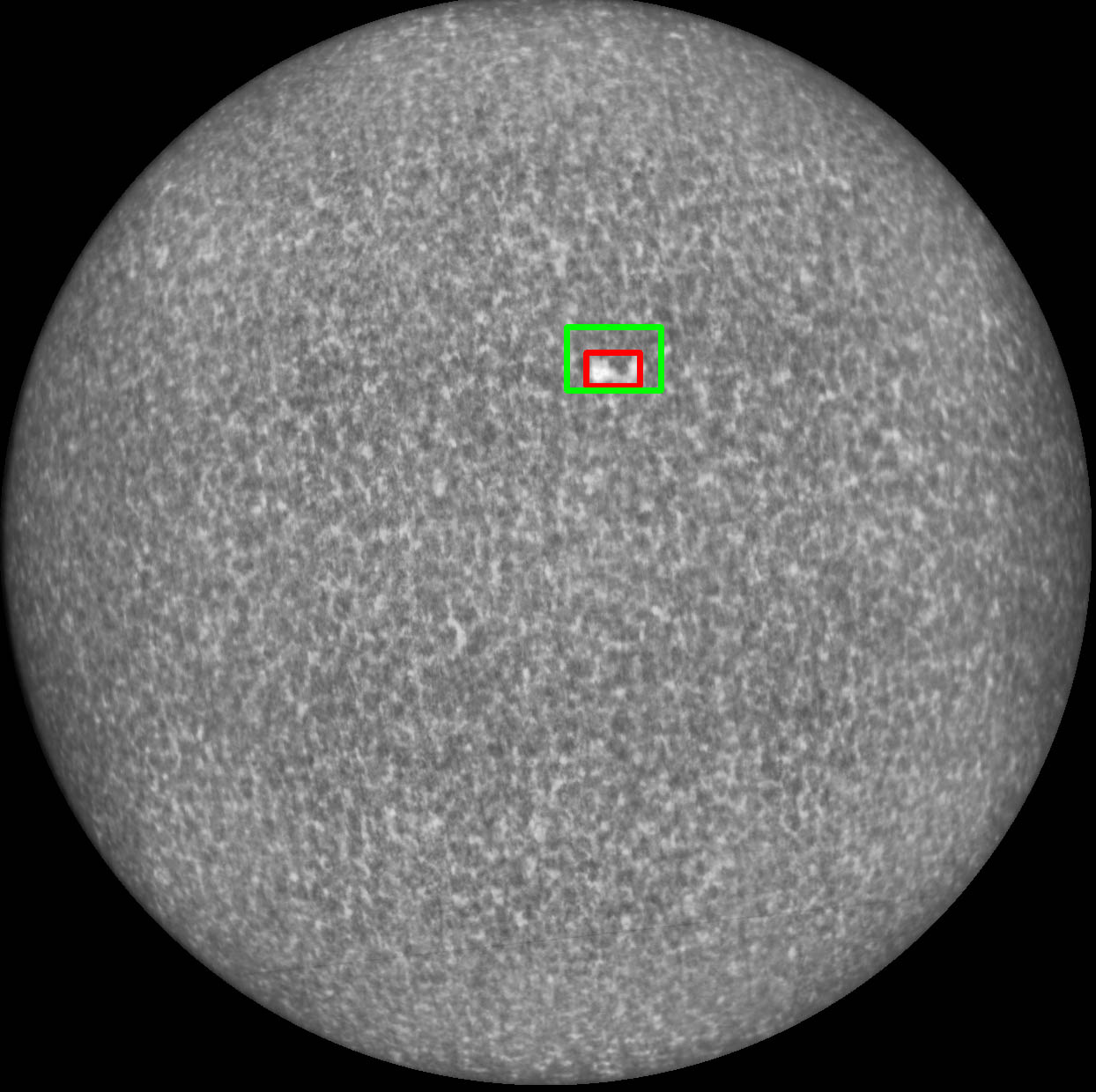} \\

\end{tabular}

\caption{Ground-truth (red) and MLMT-CNN's (green) detection of ARs at three levels of solar activity (left to right: high, medium, low) in randomly selected images from (top to bottom) SOHO/MDI Magnetogram and PM/SH 3934~{\AA}. 
}
\label{fig:LAD_activity_levels}
 \end{figure}

\begin{figure}[t]
\centering
\begin{tabular}{ccc}

\includegraphics[width=0.305\linewidth]{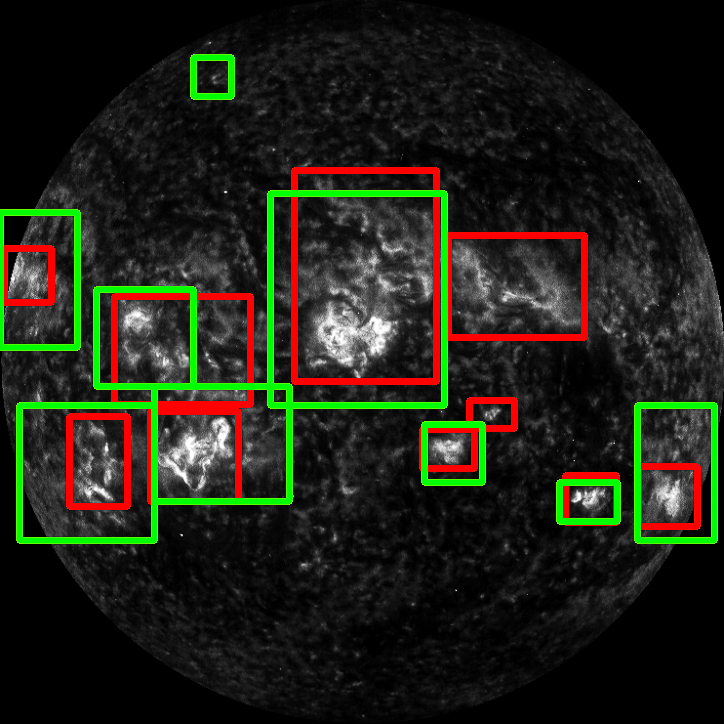} 
\includegraphics[width=0.305\linewidth]{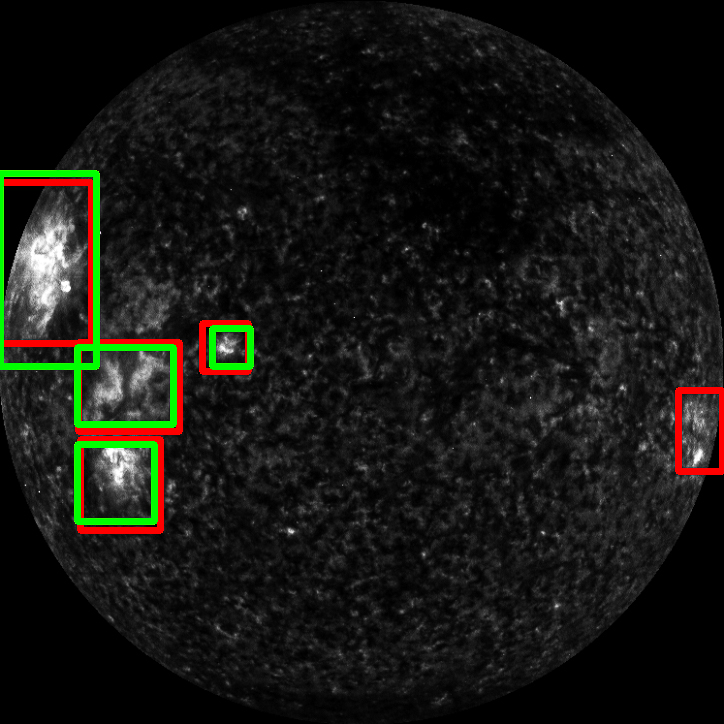} 
\includegraphics[width=0.305\linewidth]{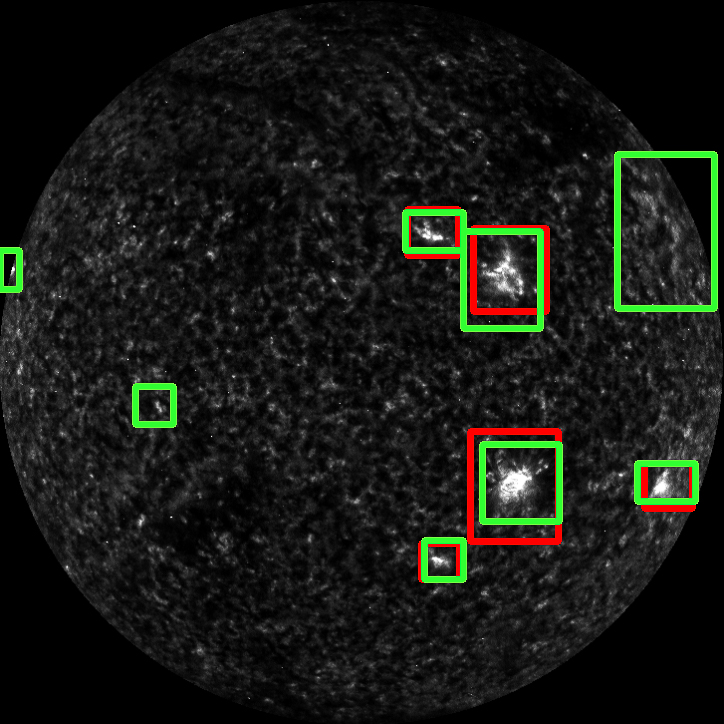} \\

\includegraphics[width=0.305\linewidth]{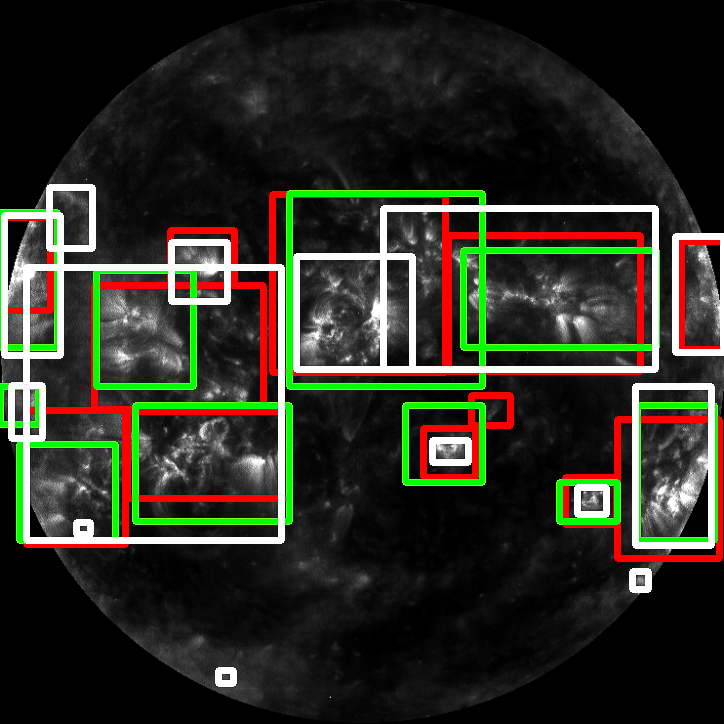} 
\includegraphics[width=0.305\linewidth]{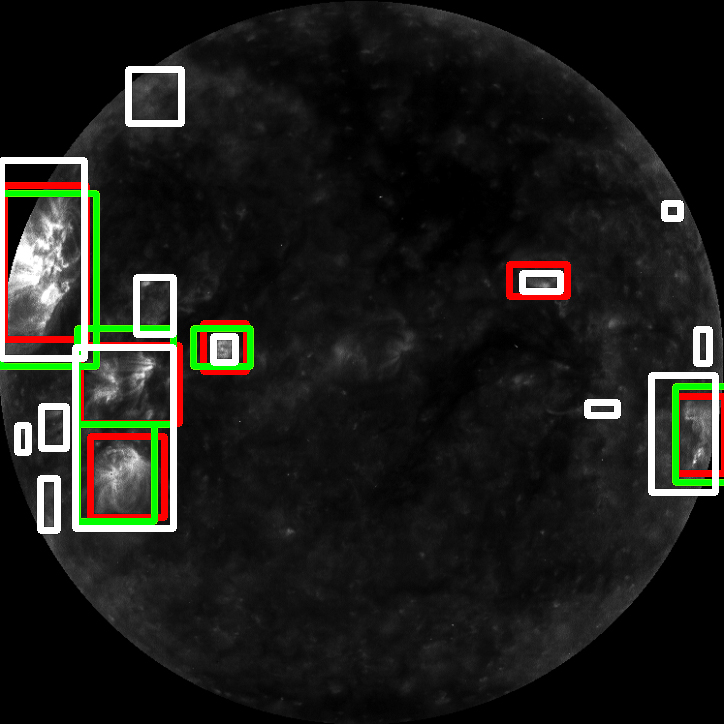} 
\includegraphics[width=0.305\linewidth]{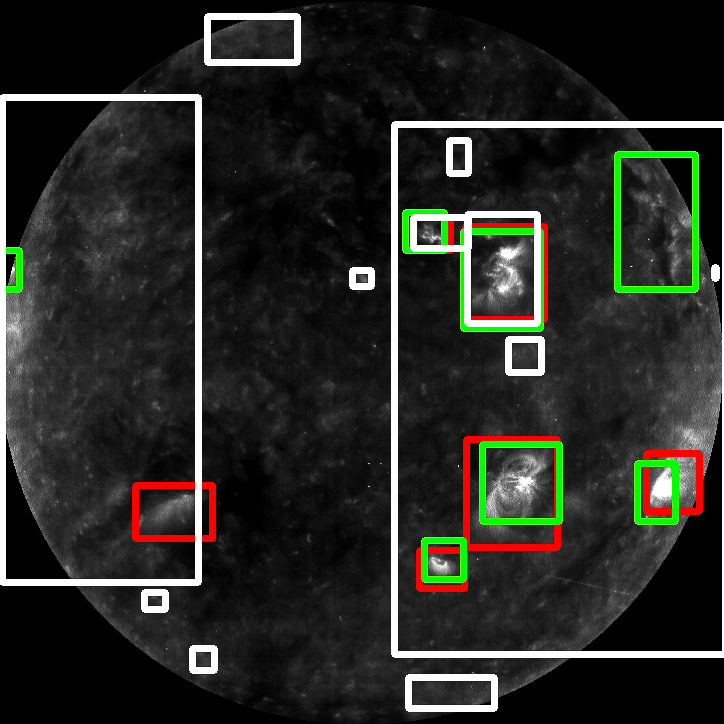} \\

\includegraphics[width=0.305\linewidth]{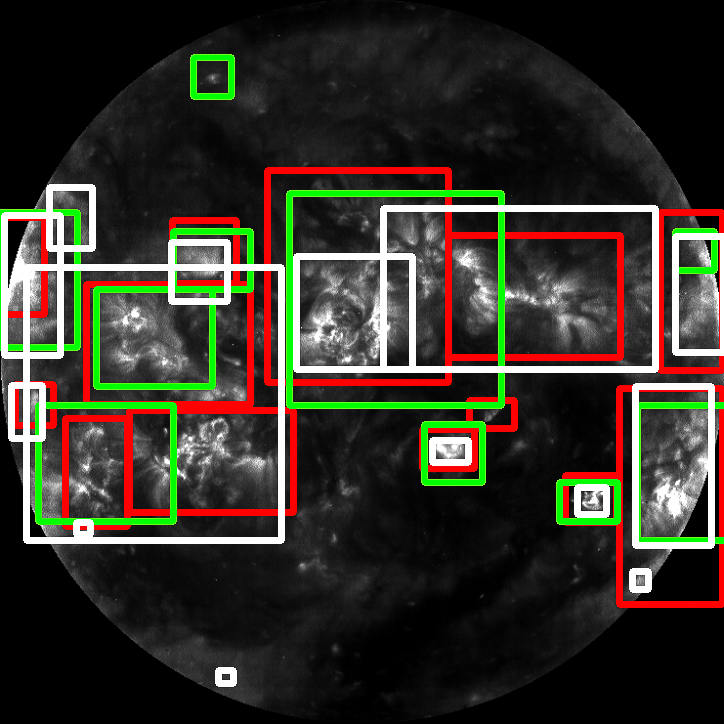} 
\includegraphics[width=0.305\linewidth]{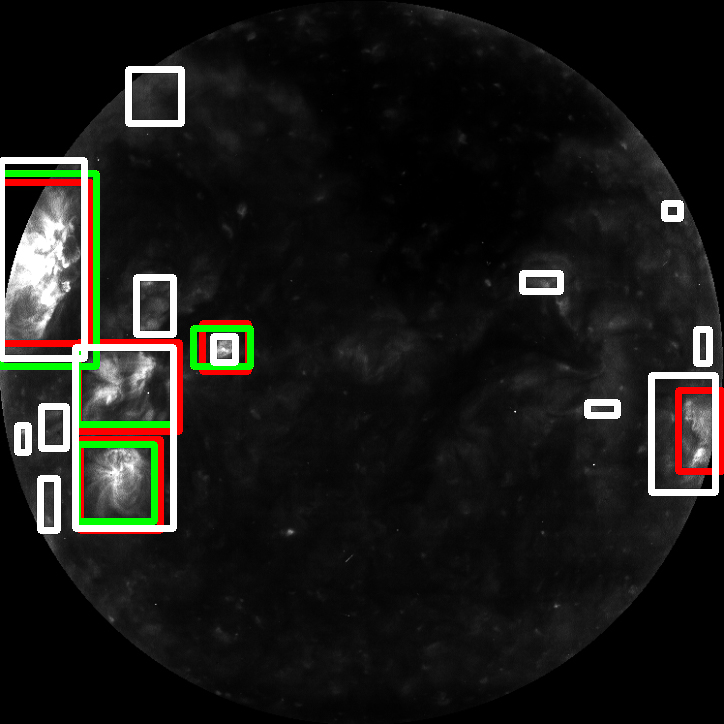} 
\includegraphics[width=0.305\linewidth]{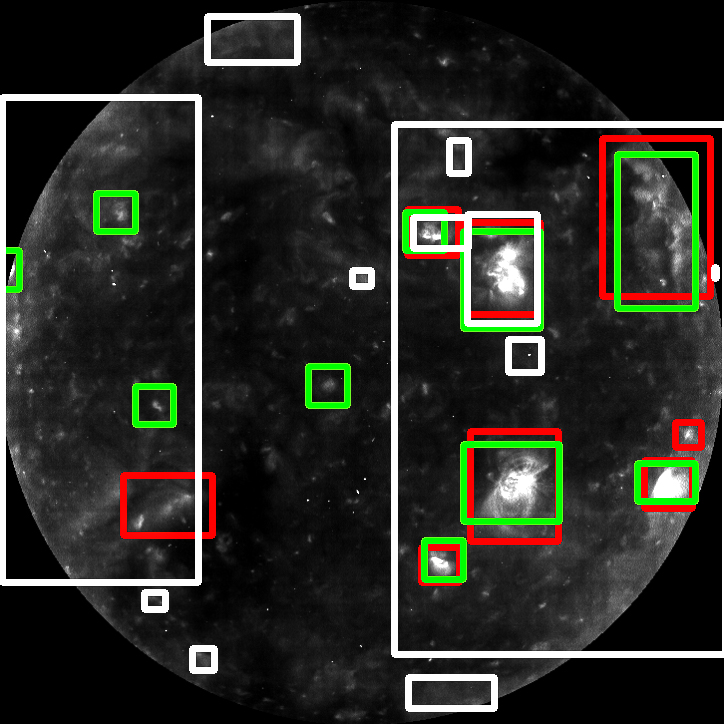} \\

\includegraphics[width=0.305\linewidth]{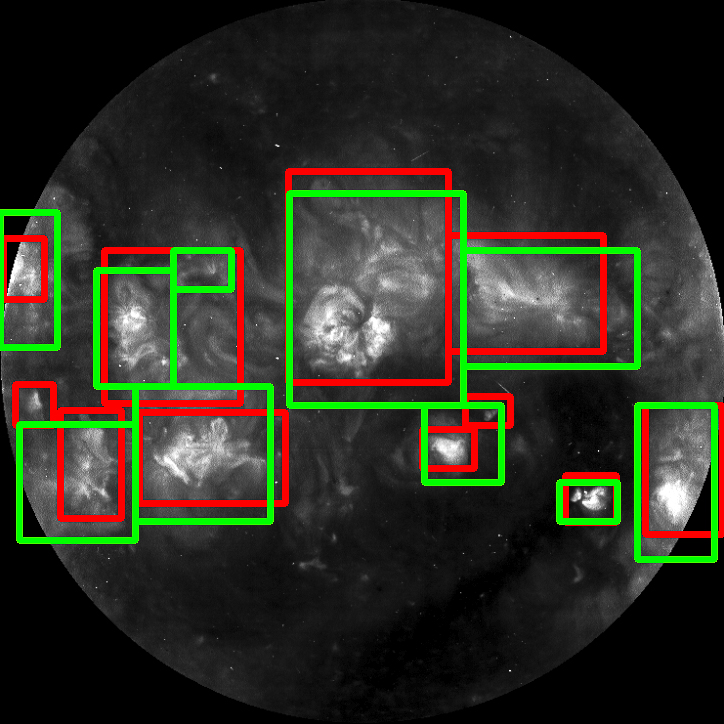} 
\includegraphics[width=0.305\linewidth]{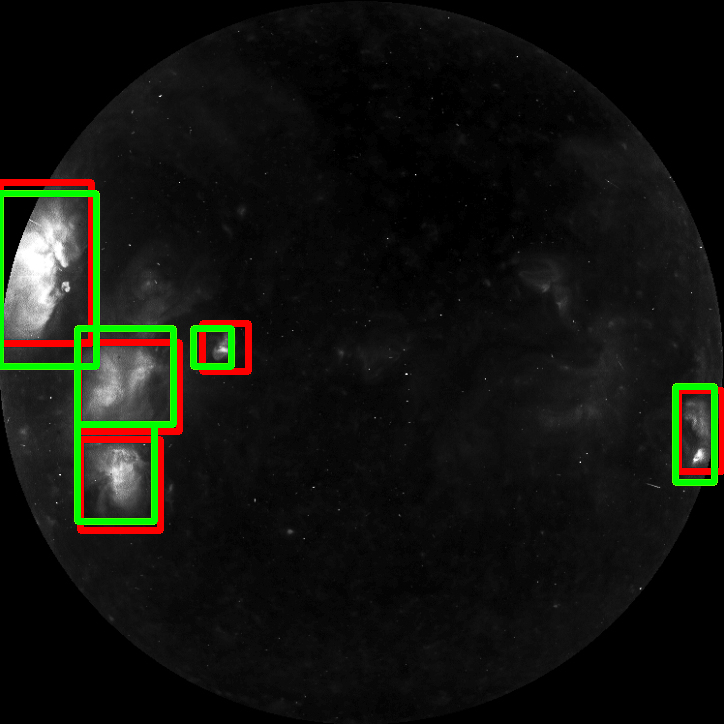} 
\includegraphics[width=0.305\linewidth]{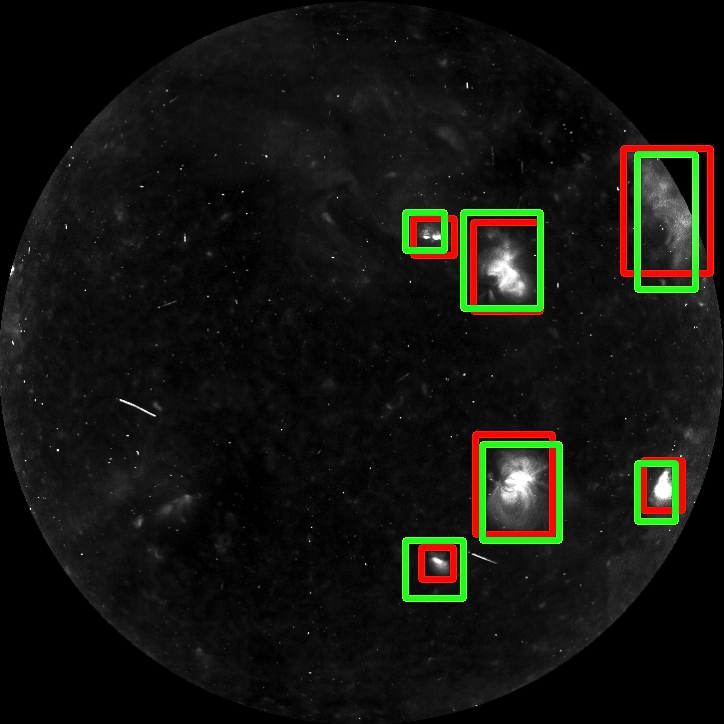} \\

\end{tabular}

\caption{
Ground-truth (red) and MLMT-CNN (green) and SPOCA’s (white) detection of ARs at three levels of solar activity (left to right: high, medium, low) in randomly selected images from (top to bottom) SOHO/EIT 304~{\AA}, 171~{\AA}, 195~{\AA}, and 284~{\AA}. 
}
\label{fig:UAD_activity_levels}
\end{figure}

Solar features (e.g. active regions (ARs)) detection and segmentation are essential in studying solar weather and behaviours. This analysis can be carried out by remotely monitoring the solar atmosphere continuously on multiple wavelengths, e.g. as shown in Figs.~\ref{fig:LAD_activity_levels} and \ref{fig:UAD_activity_levels},
captured from different ground- and space-based sensors. 

However, unlike traditional multi-spectral scenarios such as Earth imaging from space, e.g. \cite{method:ACFTHOG,method:pedestrianDet,mohajerani2019cloudnet,Mohajerani_2018,method:powerplants,method:DoDeep,method:MulVehicle}, where multiple imaging bands reveal different aspects (e.g. composition) of a same scene, in solar physics, different bands capture the solar atmosphere at different temperatures, which correspond to different altitudes \cite{paper:Fractal-Fuzzy-segmentation}.

Indeed, the solar atmosphere consists of various atoms, each of which emits light of a certain wavelength when they reach a specific temperature, in a context of strong temperature gradient across the solar atmosphere. Therefore, different wavelengths show different 2D layers of the 3D objects (e.g. ARs) that span the solar atmosphere. {We refer to this scenario as \textit{multi-layer analysis}.} For this reason, handling the multi-spectral {(and multi-layer)} nature of the problem is not straightforward.
Moreover, the variety in shapes, fuzzy boundaries, and differing brightness of ARs also make their precise localisation complex.

Very few solutions were presented to the AR localisation problem. Most of these methods exploited single image bands only, e.g. \cite{benkhalil2006active,paper:Fractal-Fuzzy-segmentation}. 
Authors justified this by the fact that each band provides information from a different solar altitude,
they show how areas of ARs differ from band to band \cite{paper:Fractal-Fuzzy-segmentation}. We , however, argue that inter-dependencies exist between bands, which can be exploited for increased robustness.

{The SPOCA method \cite{method:spoca} used clustering to extract (pixel-wise) ARs and coronal holes from SOHO/EIT 171~{\AA} and 195~{\AA} combined images,
assuming that they should yield identical detection. This approximation may result in a poor analysis of at least one of these bands.
SPOCA's detection is based on Fuzzy C-means and Possibilistic C-means \cite{Krishnapuram1996}, followed by post-processing with morphological operations. The use of fuzzy logic in 
SPOCA addresses the uncertainty in defining AR boundaries \cite{method:spoca}. The quality of results was subjectively evaluated on 112 observations. SPOCA is now used in the HFC online catalogue.
}

Generally, these methods are mainly based on clustering and morphological operations, thus are pre- and post-processing dependant, which makes them difficult to adapt to new image domains and hyperparameter-dependant. 

In this work,
we investigate the possibilities offered by deep learning (DL) methods and exploit more bands than previous methods, for richer information on the solar atmosphere.
In the past two decades, object detection has evolved dramatically,
from hand-crafted features based detection (e.g. Haar \cite{paper:violaandjones}, and HOG \cite{paper:hogfeatures}) to deep neural networks (DNN) such as YOLO \cite{method:yolo}, SSD \cite{method:SSD}, R-FCN \cite{paper:RFCN}, Cornernet \cite{paper:cornernet}, or Faster RCNN \cite{ren2015FasterRCNN}. Generally, DL based detectors rely on convolutional neural networks (CNN) to analyse images. 

{These may be split into two categories, 1) two stage detection, in which images are analysed in two steps, region proposal (generate a set of suspicious locations) and a final classification stage, and 2) one stage detection, where a DNN learns to regress object locations and classes in a single step.
In general, two stage detectors (e.g. Faster RCNN) can achieve higher accuracy over single stage detectors \cite{paper:opttradeoff, Huang_2017}.
However, such methods aim at analysing 2D images or dense 3D volumes, and are therefore not suited
to directly handle the sparse 3D nature of the solar imaging data.
Hence, we design a specialised DL framework that can
accommodate 
for different DL architectures as a backbone. We demonstrate this by applying our framework to different backbones (Faster RCNN and U-Net) and tasks (object detection and segmentation)}

Multi-spectral images are commonly treated in a similar fashion to RGB images, by stacking different bands into multi-channel images, 
\cite{method:DoDeep,method:powerplants,paper:ganimultispectral,mohajerani2019cloudnet,paper:DL-MulModal-medical-Seg}.
These methods are designed under the assumption that the different image bands capture different aspects of the same scene, which makes it ill-suited for our multi-layer case, where spatial positioning indeed differs from band to band.
Another common approach is to aggregate information from different bands at different levels (e.g. feature level and image level) \cite{paper:simonyantwo,paper:eitel_RGBD,method:MulVehicle,method:pedestrianDet,method:ACFTHOG,paper:DL-MulModal-medical-Seg,paper:mulspectpeddet}. This feature fusion strategy demonstrates the potential for DNNs to improve localisation by exploiting the multi-spectral aspect of the data.
Some works found that feature level fusion assists CNNs in producing a more consistent detection than using image level fusion for pedestrian detection from RGB and thermal images \cite{method:pedestrianDet}.
Contrary, image fusion worked best when segmenting soft tissue sarcomas in multi-modal medical images \cite{paper:DL-MulModal-medical-Seg}.
This suggests that there is no universal best fusion strategy. Thus, we investigate different types of fusion and different stages to apply fusion.
Another feature fusion strategy was used to segment coronal holes from SDO's 7 EUV bands and line-of-site magnetogram in \cite{paper:mul-ch-coronal-hole-det}. The method relies on training a CNN, using weak labels, to segment coronal holes from a single band, followed by fine-tuning the learned CNN over the other bands consecutively.
The feature maps of each specialised CNN are used in combination as input to a final segmentation CNN, resulting in a unique final prediction. 
This unique localisation result for all multi-spectral images is a common limitation to all cited works for our multi-layer scenario, which we address in this study with a multi-task network.

In this work, We introduce a novel MultiLayer MultiTask CNN (MLMT-CNN), a multi-tasking DNN framework, as a robust solution for the solar AR localisation problem (i.e. detection and segmentation) that takes into consideration the multi-layer aspect of the data and the 3-dimensional spatial dependencies between image bands. In a preliminary work \cite{method:ARDetMS}, we demonstrated its potential of analysing multiple layers simultaneously for AR detection in the form of bounding box. In this paper, we extend on this work, applying the MLMT-CNN framework to new tasks (segmentation) and to new datasets of different types, using new DNN backbones.

The 3D nature of our multi-spectral and multi-layer imaging scenario, which differs from other multi-spectral cases such as Earth observations, requires a new benchmark.
Therefore, we introduce two annotated datasets comprised of images of the solar atmosphere from both ground and space-based sensors.
They cover evenly all phases of solar activity, which follows an 11-year cycle.
To the best of our knowledge, no localisation ground-truth was previously available for such data. 
A labelling tool was hence designed to cope with its temporal, multi-spectral, and multi-layer nature and will be also released. The solar data with bounding box labels were first presented in our preliminary work \cite{method:ARDetMS}. Here, we further extend the datasets with additional weak segmentation labels.

Furthermore, we propose a training approach that accounts to the different objectives of the individual MLMT components using their correspondent losses, in contrast to the classical training in which all components are deemed to reach an optimal solution simultaneously according to their overall loss.

Our contributions may be summarised as:

\begin{enumerate}

\item We present a paradigm to handle multi-spectral solar images that show several layers of a 3D object that span the solar atmosphere (i.e. multi-layer).

\item We demonstrate the effectiveness of our approach in MLMT, a multi-task DL framework for solar AR localisation. Localisation includes both detection in the form of bounding boxes, and pixel-wise segmentation. 
We further explore and demonstrate the potential of our proposed paradigm by implementing it with different state-of-the-art CNN backbones, as well as handling different data types and arbitrary number of bands.

\item 
We propose a training strategy for MLMT that optimises the DNN weights more effectively for each objective than the classical training strategy.

\item To address the difficulty of producing accurate and detailed annotations for AR segmentation, we propose a recursive training approach based on weak labels (i.e. bounding boxes).

\item
We introduce two balanced and annotated datasets of multi-layer images of the solar atmosphere for AR detection, from both ground- and space-based data.

\item
We release a multi-spectral and multi-layer image annotation tool that facilitates bounding box labelling using temporal and spectral information.

\item We further validate our approach on an artificially created dataset of multi-modal medical images of similar spatial configurations to the multi-layer solar images.

\end{enumerate}

\section{Methodology}
\label{sec:method}

Our framework exploits several time-matched multi-layer images in parallel, to predict separate, although related, localisation results for each image.
These time-matched observations are possibly acquired by different instruments or at different orientations of the same instrument. As such they are spatially aligned prior to analysis.
Our localisation involves two stages: detection, in the form of bounding box around an object and its classification of object type, followed by a segmentation stage to produce a pixel-wise classification map enclosed in the predicted bounding box.

For both stages, we deploy a new multi-layer and multi-task DL framework that analyses information from neighbouring layers (i.e. image bands). The network learns band-specific features, these features are then fused 
at multiple levels in the network, inducing the network to learn correlations between the different bands. Finally, the resulting embeddings are jointly analysed, exploiting information from neighbouring layers to produce their separate but related results.

This framework is general and may be used with various DNN backbones.
We experiment with Faster RCNN and U-Net backbones, for detection and segmentation respectively, demonstrating the benefits of our joint analysis scheme in learning the inter-dependencies between the different image bands in both stages. Our framework may be easily adopted to serve other applications, as demonstrated with BraTS-prime and Cloud-38-prime, cf. Section \ref{sec:experiments}.

In this section, 
we introduce the main concepts of the MLMT-CNN framework in Section \ref{subsect:MSMT},
the backbone networks
used in our framework in Section \ref{sub:backbone},
and the details of our two detection and segmentation stages in Sections \ref{sub:SR-CNN-Det} and \ref{sub:SR-CNN-Seg}, respectively.

\subsection{MultiLayer-MultiTask (MLMT) framework}
\label{subsect:MSMT}

While some existing works were developed for analysing multi-spectral images, to our best knowledge, the problem of detecting objects over multi-layer imagery, which is a sparse 3D multi-spectral case in which different bands show different scenes (i.e. layers), was not yet addressed. We introduce a new multi-layer and multi-task framework (MLMT) to tackle this scenario.
The intuition behind our framework manifests in three key principles:
\begin{enumerate}
    \item Extracting features from different image bands individually using parallel feature extraction branches. This allows the network to learn independent features from each band, according to their specific modality.

    \item Aggregating the learned features from the different branches using some appropriate fusion operator. This assists the network to jointly analyse the extracted features from different bands and thus learn interdependencies between the image bands. 
    In this work, we test fusion by addition and concatenation, at different feature levels (i.e. early and late fusion).
    
    \item Generating a set of results per image band, based on a multi-task loss, allowing the detection of different sections or layers of 3D objects within the different bands. 

\end{enumerate}

Points 1 and 3 are motivated by the nature of the multi-layer data, where different bands capture different locations in a 3D scene, each providing some unique information. Our multi-tasking framework aims at obtaining specialised results for each image band, in contrast to most existing works where focus is on producing an independent prediction to all image bands. This is crucial since the localisation information may differ from one band to another in cases of multi-layer images (e.g. solar data).
Yet, all bands are correlated, which motivates point 2. Our framework exploits the inter-dependencies between the different bands by its joint analysis strategy, increasing the robustness of its performance in individual bands.

Furthermore, our framework emulates how experts manually detect ARs, where a suspected region's correlation with other bands is evaluated prior to its final classification. This demonstrates the usefulness and importance of accounting for (spatially and temporally) neighbouring slices in robustly detecting ARs.

Moreover, this framework is very modular and flexible. It can accommodate any number of available image bands (i.e. layers) and perform different tasks (e.g. detection and segmentation). Additionally, since different scenarios may require different fusion strategies (as suggested by existing works), the modularity of our framework allows it to be easily adapted to different cases. We demonstrate this by applying our framework to different applications in Section \ref{sec:experiments} (solar ARs, BraTS-prime, and Cloud-38-prime datasets), where we investigate the best type and level of feature fusion (e.g. addition and concatenation, early and late).

\subsection{Backbone networks}
\label{sub:backbone}

The modular design of our framework allow it to adopt different backbone architectures. Indeed, the 3 key principles are applicable to different backbones, as they are not architecture dependent. We demonstrate this in this section and discuss different backbone networks for different tasks in which we adopt our framework to.

\subsubsection{Detection backbone: Faster RCNN}
\label{sub:faster-r-cnn}

For detection, we adopt the Faster RCNN architecture as the backbone.
Faster RCNN is a DL-based detector that may be trained to detect and classify a number of objects from a (usually RGB) image. It consists of three main parts: 1) convolutional layers extract features from the input image, as in any CNN. From these features, 2) a region proposal network (RPN) proposes locations that might contain objects, and 3) a detection network predicts the object class of each proposed locations. We apply our framework to the three stages detection strategy of Faster RCNN, thus generalising it to jointly analysing multiple images that span different locations (or layers) of a same 3D scene.

Comparing to other state-of-the-art architectures (e.g. YOLO and SSD), the multi-stage design of Faster RCNN allows aggregating information from different bands at different levels, namely low level (i.e. feature extraction stage) and high level information (i.e. region proposals). Additionally, Faster RCNN has scored the highest accuracy in \cite{Huang_2017}.

\subsubsection{Segmentation backbone: U-Net}
\label{subsubsec:unet}

We experiment with U-Net as the backbone of our segmentation stage. Nevertheless, other competing networks can also be used, and we also experimented with FCN8 \cite{paper:FCN8} in early tests.
U-Net \cite{paper:UNet} is a fully convolutional network that consists of 3 main parts: 1) contraction path, 2) bottleneck, and 3) expansion path. 

In our segmentation stage, we apply our MLMT framework to the building blocks from U-Net to demonstrate the benefits of the joint analysis in segmenting ARs.
MLMT takes advantage of U-Net's skip connections that allow combining features from different semantic levels within the same band. This maximises the learned information within individual bands while combining this information with feature fusion at the U-Net's bottleneck stage.
Thus, information from different bands are combined for classification, while preserving the spatial information of individual images.

\subsection{MLMT-CNN: Detection stage}
\label{sub:SR-CNN-Det}
\label{subsub:-Det-approach}

\begin{figure}
	\centering
		\includegraphics[width= \linewidth]{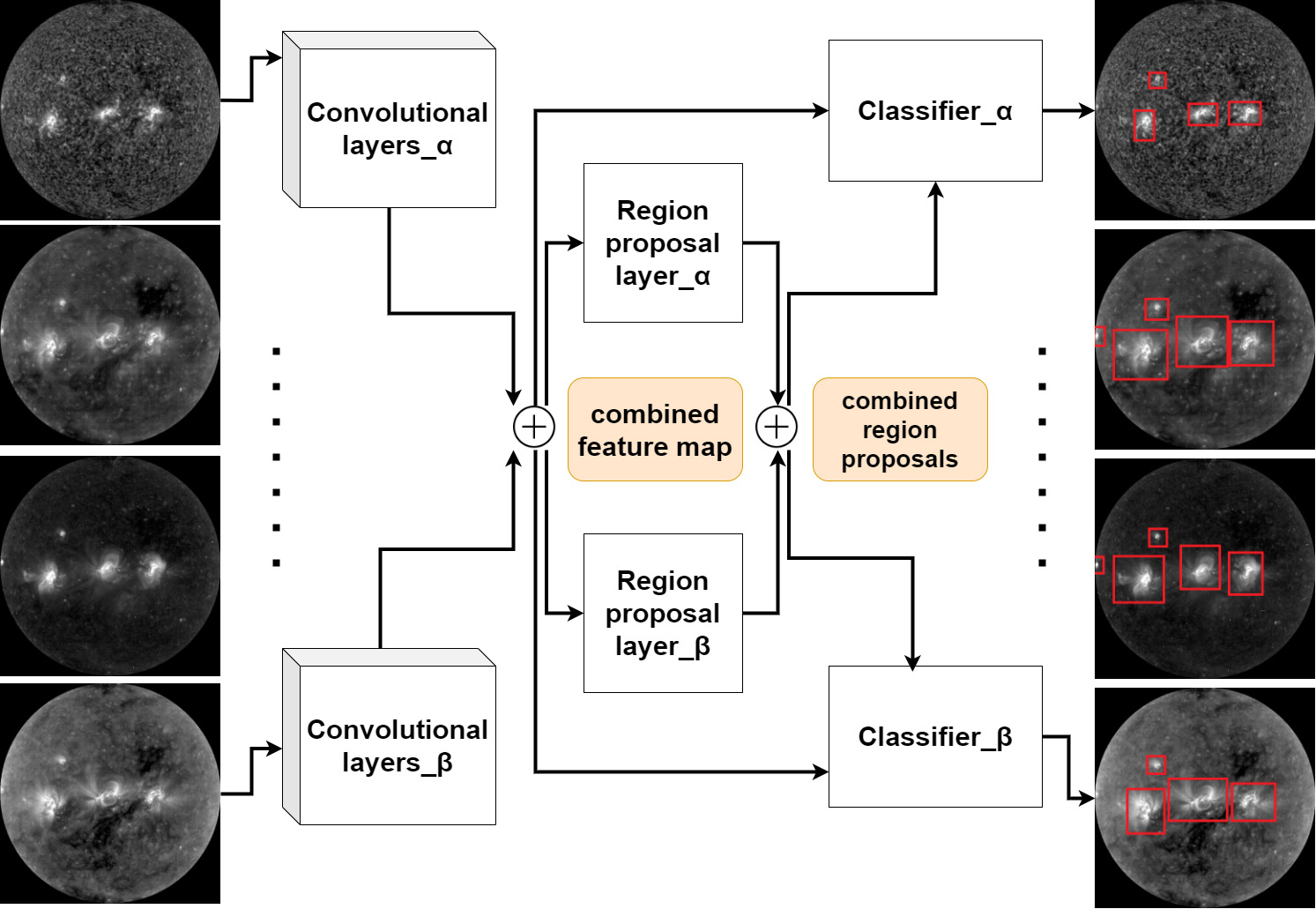}
	\caption{MLMT for detection using the Faster RCNN backbone. The '{$+$}' sign denotes concatenation of the feature maps, or of the lists of region proposals.
	}
	\label{fig: Spectral-R-CNN architecture and flow of processes}
\end{figure}

Our detection DNN is presented in Fig.~\ref{fig: Spectral-R-CNN architecture and flow of processes}. It takes the pre-processed multi-layer image as input.
A CNN (ResNet50 or VGG16 in our experiments) is first used as a feature extraction network. Parallel branches (subnetworks) produce a feature map per image band, following the late (or feature map) fusion strategy.
Since individual bands provide different information, this allows the subnetworks' filters to be optimised for their input bands individually. The feature maps are then concatenated across the bands, assisting the network to learn correlations between them.

The combined feature map is jointly analysed by one parallel module per image band that performs Faster RCNN's RPN. The RPN stage uses three aspect ratios ([1:1], [1:2], [2:1]) and four sizes of anchor (32, 64, 128, and 256 pixel width). We found empirically that these match well the typical size and shape of ARs. One specialised RPN per image band is trained. 

At training, for each band, the correspondent region proposals along with the combined feature map are used by a detection module to perform the final prediction for the band. However, at testing time, the band-specialised detector modules use the region proposals from all bands.
This combination of region proposals helps finding potential AR locations (i.e. region proposals) in bands where they are more difficult to identify. This aids the network to learn the correlation between the different bands more dynamically, benefiting from information from different bands simultaneously while having band-specialised region proposal and detection models.

It is worth noting that during training, the RPN proposals for a band are filtered (i.e. labelled as positive or negative) with respect to their overlap with the band's own ground-truth. Hence, combining them in the training time would mean implicitly inheriting the ground-truth of a band to another, in contradiction with the band-specific ground-truth used for training the detector module. Indeed, different bands show distinct cuts of a 3D object in which each cut must have its own ground-truth. Combining ground-truths of different bands at training time may hinder the learning of both the RPN and detector modules. Therefore, region proposals are only combined at testing time to ensure a better learning of the final detection modules. 

Using the combined feature map aids the network to learn the relationship between the image bands, in both region proposal and classification stages, hence providing a more robust prediction in line with the nature of the data. This prediction is still band-specialised thanks to the different ground-truths being used for each band at training time. We demonstrate in Section \ref{sec:experiments} that this is particularly helpful in cases where an AR is difficult to detect in a single band.

We train our MLMT framework using all input bands and branches according to a combined loss function:

\begin{equation}
\begin{gathered}
\resizebox{.91\hsize}{!}{
\label{loss:MSMT}
$L = \sum_{b}^{}
\Bigg(\frac{1}{N_{cls}} \sum_{i}^{ } L_{cls}(p_{b_{i}}, p^\ast_{b_i}) + \lambda \frac{1}{N_{reg}} \sum_{i}^{ }p^\ast_{b_i} L_{reg}(t_{b_i}, t^\ast_{b_i}) \Bigg) 
$}
\end{gathered}
\end{equation}

where {$b$} and {$i$} refer to the image band and the index of the bounding box being processed, respectively. The terms $L_{cls}$ and $L_{reg}$ are the bounding-box classification loss and the bounding-box regression loss defined in \cite{ren2015FasterRCNN}.
$N_{cls}$ and $N_{reg}$ represent the size of the mini batch being processed and the number of anchors, respectively. $\lambda$ balances the classification and the regression losses (we set $\lambda$ to 10 as suggested in \cite{ren2015FasterRCNN}). $p$ and $p^\star$ are the predicted anchor's class probability and its actual label, respectively. Lastly, $t$ and $t^\star$ represent the predicted bounding box coordinates and the ground-truth coordinates, respectively.
It is worth noting that our proposed framework is not limited to using Faster RCNN's loss and may be trained with using other task-suitable loss functions.

During training, the weights of each stage (i.e. feature extraction, region proposal, and detection) are stored independently whenever the related Faster RCNN loss decreases. At testing time, the best performing set of weights is retrieved per stage.
We refer to this practice as `Multi-Objective Optimisation' (MOO). The improved performance that we observe in Section \ref{sec:experiments} may be explained by each stage having a different objective to optimise, which may be reached at different times.

In this paper, we experiment with a 2, 3, and 4-band pipeline. However, the approach may generalise straightforwardly to $n$ bands and new imaging modalities.

\subsection{MLMT-CNN: Segmentation stage}
\label{sub:SR-CNN-Seg}

\begin{figure}
	\centering
		\includegraphics[width=1.0\linewidth]{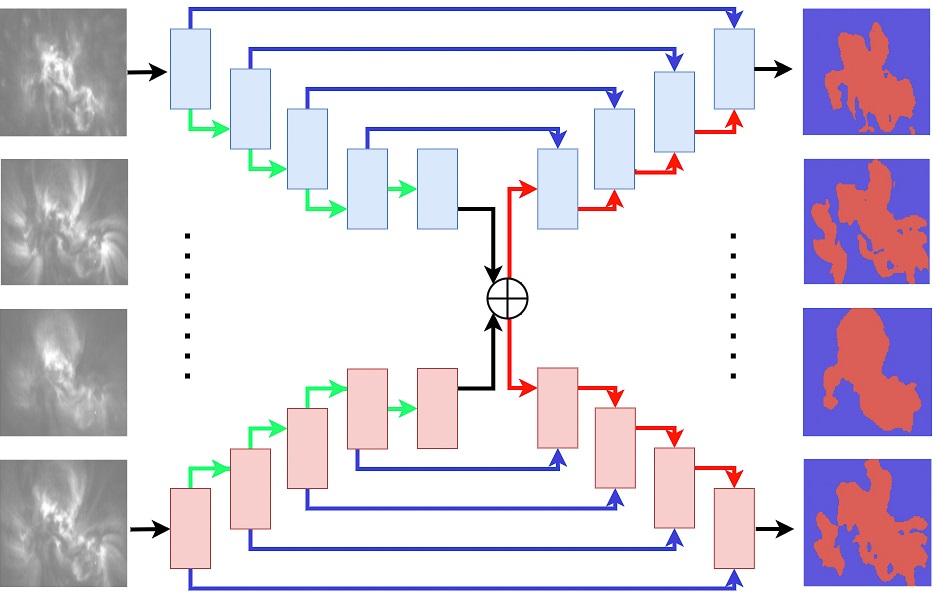}
	\caption{MLMT for segmentation using the U-Net backbone. The '{$+$}' sign denotes fusion of the feature maps. 
	Coloured boxes are convolutional blocks for each branch (band) respectively. Green and red arrows denote max pooling and up sampling operations, respectively. Blue arrows are skip connections, applied to the appropriate channel of the joint feature map for each branch.}
	\label{fig:Solar_SegNet_arch}
\end{figure}

Our segmentation framework is presented in Fig.~\ref{fig:Solar_SegNet_arch}. It consists of 3 parts: 1- feature extraction, 2- feature fusion, and 3- mask reconstruction.
The network takes as input the AR detections (patches) produced by the detection stage. Each detection is cropped from all image bands,
and resized into 224x224 pixel before entering the segmentation network.

The feature extraction part consists of parallel U-Net contracting paths (one per band), each specialised to extract a feature map from its band individually.
The resulting feature maps are then combined in the latent space (i.e. late fusion).
It is worth noting that different feature fusion operations may be used. In this work, we experiment with addition and concatenation.
The combined feature map is passed to the mask reconstruction part where parallel U-Net expensive paths (a specialised path per band) perform the final prediction. Skip connections are utilised between each band's contracting path and its correspondent expensive path to preserve fine details learned in early layers of that band (blue arrows in Fig. \ref{fig:Solar_SegNet_arch} ).

To overcome the lack of dense AR annotation, we use weak labels to train our segmentation network along with a recursive training approach.
In the first round of iterations, weak annotations are used to guide the training. Once the network converges, the training is repeated from random weights using the new labels predicted by the model from the previous round. This process is repeated until validation loss stops decreasing, or starts to increase. The idea is inspired by \cite{paper:simple-does-it,paper:weakly_panoptic_seg,paper:boxsup,paper:weakly-supervised-seg-remote-sensing}, where authors demonstrate that iteratively training segmentation CNNs with weak labels can achieve results close to fully supervised.

Our weak label was carefully designed to provide a conservative representation of ARs, favouring a high precision over recall, to accelerate the first training round (as detailed in \ref{sec:experiments}). Recursive training allows the network to learn a more generalised representation by supervising itself through the recursion process, while limiting the bias that may be introduced by the initial weak label. This is in line with the discovery that sampling as little as 4\% of the pixels to compute the training loss enables CNNs to achieve a close performance to fully supervised, caused by the strong correlation within the training data of a pixel-level task \cite{paper:pixelnet}.
The results of our recursive approach were validated by a solar physics expert, and will be further discussed in Section \ref{sec:experiments}.

Moreover, the solar data suffers from a class imbalance by nature, since most of the solar disk is covered by quite sun (solar background). The use of AR crops (patches from previous detection) helps in reducing this imbalance significantly, yet it does not solve the matter completely. Hence, we train our model using a weighted categorical cross entropy loss that combines information from all image bands as follows:
\begin{equation}
\label{loss:Seg}
\begin{split}
L(y, \hat{y})= - \sum_{b=0}^{} \sum_{c=0}^{} \omega_c \sum_{i=0}^{} y_{icb} ~*~ log ~(\hat{y}_{icb})
\end{split}
\end{equation}
where \textit{y} and \textit{$\hat{y}$} are the actual and the predicted classes, respectively, \textit{${\omega}_c$} is the weight of the $c^{\text{th}}$ class, and \textit{i} and \textit{b} denote the pixel and the band being processed, respectively.
We use the values 2, 1, and 2 as the weights for the three AR, solar background (quite sun), and image background classes, respectively. These weights were found to be best performing by experimenting with different values based on prior computed class ratios. Adding the weighting term to the combined loss prevents any bias that might be caused by the dominating solar background class.

\section{Experiments}
\label{sec:experiments}

All experiments were implemented using Tensorflow with an NVIDIA GeForce GTX 1080 Ti GPU. The detection and segmentation stages were trained for 3000, and 250 epochs ($\sim$4 and $\sim$0.41 days), respectively, using Adam optimiser \cite{paper:adam} with learning rates of 
2e-5 and 4e-3, respectively.

\subsection{Data}
\label{sec:data}

\subsubsection{Labelled AR datasets}

We work with images from SOHO spacecraft
and Paris-Meudon (PM) observatory.
Multi-layer solar images comprise of measurements at different ultraviolet and X-ray wavelengths (denoted as \textit{bands}) and centred on the emission wavelengths of ionised atoms of interest. Since these ionised atoms exist at given temperatures, they allow imaging different altitude regions of the solar atmosphere, following its temperature gradient.
ARs are areas of strong magnetic field. Therefore, the multi-spectral and multi-layer images may be complemented by magnetograms that inform on the intensity and polarity of the magnetic field. With current technologies, magnetograms are mainly available for the photosphere.
The images of this study were acquired in the 171~{\AA}, {195~{\AA}}, 284~{\AA}, and 304~{\AA} bands (SOHO/EIT imager),
3934~{\AA} band (PM Spectroheliograph (PM/SH) imager),
and the magnetogram images (SOHO/MDI imager)
as illustrated in Figs.~\ref{fig:LAD_activity_levels} and \ref{fig:UAD_activity_levels}. These correspond to observing the photosphere (magnetogram), chromosphere (3934~{\AA}), chromosphere and base of the transition region (304~{\AA}), transition region (171~{\AA} {and 195~{\AA}}), and corona (284~{\AA}). Solar observations are acquired frequently to study the evolution of solar features and events over time.

\label{sub:labelling}

Our work requires ground-truth annotations of ARs in the form of bounding boxes (detection) and pixel-wise masks (segmentation). 
To the best of our knowledge, no such annotated dataset is currently publicly available. Therefore, we publish 
the Lower Atmosphere Dataset (LAD) and Upper Atmosphere Dataset (UAD).
Both datasets include bounding box annotations produced with a new multi-spectral labelling tool.
which displays, side by side, images from an auxiliary modality and from a sequence of 3 previous and 3 subsequent time steps.
ARs have a high spatial coherence in 3934~{\AA} and magnetogram images due to the physical proximity of the two imaged regions, hence they share the same bounding boxes. The UAD additionally includes weak segmentation labels
produced by thresholding and morphological operations so as to label only pixels that have an evident activity, i.e. being the brightest regions in the solar disk. This is motivated by the discovery in \cite{paper:pixelnet} that training data of a pixel-level task has a strong between-sample correlation, and that randomly sampling as little as 4\% of the pixels to train a CNN can achieve about the same performance as full supervision.
Both datasets are augmented using north-south mirroring, east-west mirroring, and a combination of the two. 
All annotations were validated by a solar physics expert.

{We split the datasets into training and testing sets in the following proportions. For LAD, we use 213 images (1380 bounding box) for training, and 53 images (406 bounding box) for testing. For UAD, we use 283 images for training, and 40 images for testing. This amounts to 2205, 1919, 2341, and 2016 training bounding boxes in the 304~{\AA}, 171~{\AA}, 195~{\AA}, and 284~{\AA} bands respectively, and 287, 262, 330, and 263 testing bounding boxes.
Furthermore, in order to compare against the localisation of SPOCA, we consider a subset of the UAD testing set for which SPOCA detection results are available in HFC: the SPOCA subset. It consists of 26 testing images (181, 168, 213, and 166 bounding boxes in the 304~{\AA}, 171~{\AA},  195~{\AA} and 284~{\AA} images respectively).}

\subsubsection{Weak-BraTS-prime}
\label{subsection:brats-prime}

To further demonstrate the benefits of our joint analysis based approach, 
we create a synthetic dataset from the BraTS 
multi-modal dataset \cite{BraTS:1} of similar spatial configurations to the solar imaging bands.
BraTS consists of full 3D MR image volumes of brain in 4 modalities (T1GD, T1, T2, and Flair) and 3 classes: enhancing tumour (ET), necrotic and non-enhancing tumour core (NCR/NET), and peritumoural edema (ED). We create the synthetic dataset by selecting one 2D slice of each image modality
separated by (spatial) gaps of size $g$. This emulates the solar images scenario where each band shows ARs in a different solar altitude. We experiment with $g$ being either 1, 2, or 3 voxels, to show the influence of the image modalities having different levels of spatial correlation on the segmentation. 
For each modality, we use a total of 11,533 and 190 training and testing images, respectively.

\subsubsection{Weak-Cloud-38}
\label{subsubsection:weak-cloud-38}

We further evaluate our recursive training approach on a third weakly labelled 
dataset derived from the Cloud-38 \cite{Mohajerani_2018} multi-modal (4 bands) dataset.
This dataset has resemblance to our solar images in that there are a variety of cloud shapes, sizes and densities, albeit the multi-layer (3D) aspect is missing.
It consists of 2,502 (2,382 training and 120 testing) images per band. 
We augment the training set using similar transformations to solar images.

\subsection{Detection stage}
\label{sub:detection-stage-results}

A detection is considered a true positive if its intersection with a ground-truth box is greater or equal to $50\%$ of either the predicted or ground-truth area. 
NMS is used in all experiments to discard any redundant detections.

All tested deep learning architectures were initialised with a pre-trained CNN with ImageNet weights (similar transfer learning strategy has been found useful in, for instance, depth estimation \cite{Crabbe2015}). 
Its worth noting that the components of each detection branch (feature extraction network, RPN, and detection network) adopt a similar hyper-parameter configuration to that suggested in Faster RCNN \cite{ren2015FasterRCNN}.

A single-channel solar image was repeated along the depth axis resulting in a 3-channel image matching the pre-trained CNN's input depth.

HFC's SPOCA detections were obtained from 171~{\AA} and 195~{\AA} images only, combined as two channels of an RGB image, and SPOCA produces a single detection for both bands. We compare this detection against the ground-truth detections of each of the bands, individually.
SPOCA may only combine image bands that are located close to each other in the solar atmosphere and for which it makes sense to produce a common set of detection results. Thus, HFC's SPOCA results are only available for bands of the transition region (171~{\AA}) and low corona (195~{\AA}), and no images from the chromosphere (304~{\AA}) or the high corona (284~{\AA}) were used.  However, to prove the robustness and versatility of our detector, we also experiment with a combination of chromosphere, transition region, and corona bands on the SPOCA subset in addition to the whole UAD.

\subsubsection{Independent detection on single image bands}
\label{sub:single-band-results}

We first compare detection results produced by Faster RCNN over individual image bands
(Table \ref{table:single-band-raw_images}). This
serves as baseline to assess our proposed framework.
Different DL-based feature extraction networks are tested (ResNet50 and VGG), 
and we present here results of the best performing, namely ResNet50.

\begin{table}
\centering
\caption{Baseline detection performance of the single image band detectors.}
\label{table:single-band-raw_images}
\resizebox{!}{.06\paperheight}{
\begin{tabular}{ccccc}

\hline
Dataset & Image band & Precision & Recall  & F1  \\

\hline
\multirow{2}{*}{LAD} &  {3934 {\AA}} & {{0.93}} & {{0.82}} & {{0.87}} \\
  &  {Magn. } & {{0.89}} & {{0.78}} & {{0.83}} \\
  \hline
 \multirow{4}{*}{UAD} &   {304 {\AA}}   & {{0.73}}  & {{0.83}} &  {{0.78}}   \\
  &   {171 {\AA}}   & {{0.84}}  & {{0.89}} &  {{0.86}}   \\
  &   {195 {\AA}}   & {{0.81}}  & {{0.75}} &  {{0.78}}   \\
  &   {284 {\AA}}   & {{0.86}} & {{0.82}} &  {{0.84}} \\
\hline

 \multirow{4}{*}{SPOCA} &   {304 {\AA}}   & {{0.72}}  & {{0.82}} &  {{0.77}}   \\
  &   {171 {\AA}}   & {{0.87}}  & {{0.87}} &  {{0.87}}   \\
  &   {195 {\AA}}   & {{0.82}}  & {{0.73}} &  {{0.77}}   \\
  &   {284 {\AA}}   & {{0.86}} &
{{0.82}} &  {{0.84}} \\

\hline
\end{tabular}
} 
\end{table}

\begin{table}[!ht]
\centering
\caption{Detection performance of the MLMT-CNN detectors. For each band, the highest scores are highlighted in bold.}
\label{table:eval_SRCNN_Raw_images}

\resizebox{!}{.2343\paperheight}{
 \begin{tabular}{ccccccc}

\hline
Detector & Fusion & Dataset & Bands & Prec. & Recall  & F1  \\

\hline
\multirow{6}{1.3cm}{MLMT-CNN (ResNet50 -- MOO)} & \multirow{2}{1.1cm}{Early -- concat.} & \multirow{6}{*}{LAD} &  {3934~{\AA}} & {0.96} & {\textbf{0.82}} & {\textbf{0.89}} \\
& &  &  Magn. & {0.95} & {0.82} & {0.88}  \\

\hhline{~~~----}

  & \multirow{2}{1.1cm}{Late -- concat.} &  &  {3934~{\AA}} & {\textbf{0.97}} & {\textbf{0.82}} & {\textbf{0.89}} \\
& &  &  Magn. & {\textbf{0.96}} & {\textbf{0.85}} & {\textbf{0.90}}  \\

\hhline{~~~----}

  & \multirow{2}{1.1cm}{Late -- addition} &  &  {3934~{\AA}} & {0.95} & {\textbf{0.82}} & {0.88} \\
& &  &  Magn. & {0.94} & {0.80} & {0.87}  \\

\hline
\multirow{8}{1.3cm}{MLMT-CNN (ResNet50)} & \multirow{8}{1cm}{Late -- concat.} &

\multirow{8}{*}{UAD} & 171~{\AA} &
{0.92} &
{0.77} &
{0.84}  \\
& & & 284~{\AA} &
{0.90} &
{0.81} &
{0.85} \\
\hhline{~~~----}

& & & 171~{\AA} &
{0.82} &
{0.85} &
{0.83}  \\
& & & 195~{\AA} &
{0.86} &
{0.72} &
{0.78} \\
\hhline{~~~----}

& & & 195~{\AA} &
{0.88} &
{0.67} &
{0.77}  \\
& & & 284~{\AA} &
{0.84} &
{0.78} &
{0.81} \\
\hhline{~~~----}

& & & 304~{\AA} &
{0.82} &
{0.79} &
\textbf{0.80}  \\
& & & 195~{\AA} &
{0.87} &
{0.75} &
{0.80} \\

\hline
\multirow{23}{1.3cm}{MLMT-CNN (ResNet50 -- MOO)} & \multirow{23}{1cm}{Late -- concat.} &

 \multirow{2}{*}{UAD} &  {171~{\AA}} & {0.90} &{0.83} & {\textbf{0.87}}  \\

& & & 284~{\AA} & {\textbf{0.93}} & {0.80} & {\textbf{0.86}} \\
\hhline{~~~----}

& & \multirow{2}{*}{SPOCA} & 171~{\AA} &
{\textbf{0.89}} &
{0.83} &
{\textbf{0.86}}  \\

& & & 284~{\AA} &
{\textbf{0.92}} &
{\textbf{0.80}} &
{\textbf{0.86}} \\
\hhline{~~~----}

& & \multirow{2}{*}{UAD} &  {171~{\AA}} & {0.86} &{0.77} & {0.82}  \\

& & & 195~{\AA} & {0.89} & {0.75} & {0.81} \\
\hhline{~~~----}

& & \multirow{2}{*}{SPOCA} & 171~{\AA} &
{0.83} &
{0.77} &
{0.80}  \\

& & & 195~{\AA} &
{0.86} &
{0.73} &
{0.79} \\
\hhline{~~~----}

& & \multirow{2}{*}{UAD} &  {195~{\AA}} & {0.88} &{0.68} & {0.77}  \\

& & & 284~{\AA} & {0.84} & {0.78} & {0.81} \\
\hhline{~~~----}

& & \multirow{2}{*}{SPOCA} & 195~{\AA} &
{\textbf{0.87}} &
{0.67} &
{0.75}  \\

& & & 284~{\AA} &
{0.81} &
{0.78} &
{0.80} \\
\hhline{~~~----}

& & \multirow{2}{*}{UAD} & 304~{\AA} & {0.82} & {0.78} & {\textbf{0.80}} \\

& &  &  195~{\AA} & {0.88} & {\textbf{0.78}} & {\textbf{0.83}} \\
\hhline{~~~----}

& & \multirow{2}{*}{SPOCA} & 304~{\AA} &
{\textbf{0.79}} &
{\textbf{0.78}} &
{\textbf{0.79}}  \\

& & & 195~{\AA} &
{0.85} &
{0.77} &
{\textbf{0.81}} \\
\hhline{~~~----}

& & \multirow{3}{*}{UAD} & 304~{\AA} & {0.78} & {0.74} & {0.76} \\

& &  &  171~{\AA} & {0.76} & {0.76} & {0.76} \\

& &  &  284~{\AA} & {0.79} & {0.78} & {0.78} \\
\hhline{~~~----}

& & \multirow{4}{*}{UAD} & 304~{\AA} & {\textbf{0.93}} & {0.69} & {0.79} \\

& &  &  171~{\AA} & {\textbf{0.94}} & {0.66} & {0.78} \\

& &  &  195~{\AA} & {\textbf{0.91}} & {0.72} & {0.80} \\

& &  &  284~{\AA} & {\textbf{0.93}} & {0.66} & {0.77} \\

\hline
\multirow{2}{*}{SPOCA}
& \multirow{2}{1.1cm}{Early -- concat.} & \multirow{2}{*}{SPOCA} &   171 {\AA}   & {0.54} & {\textbf{0.93}} &  {0.68} \\
&  & &   {195 {\AA}}   & {0.58}  & \textbf{0.82} &  {0.68}
\\

\hline

\multirow{4}{1.4cm}{\cite{paper:mul-ch-coronal-hole-det} using Faster RCNN (ResNet50)}  
& \multirow{4}{1cm}{Sequential fine-tuning} & \multirow{4}{*}{UAD} &   304 {\AA}   & {0.73} & \textbf{0.83} &  {0.78} \\
&  & &   {171 {\AA}}   & {0.80}  & {\textbf{0.90}} &  {0.84}   \\
& & &   {195 {\AA}}   & {0.83}  & {0.72} &  {0.77}   \\
& & &   {284 {\AA}}   & {0.86} & {0.80} &  {0.83} \\

\hline
\end{tabular}
} 
\end{table}

\begin{table}[htb]
\centering
\caption{F1-scores of single image band based detectors against MLMT-CNN with different fusion strategies over BraTS-prime (with 1 slice gap). All detectors are based on ResNet50. For each band, the highest scores are highlighted in bold.}
\label{table:brats_det_results_short}
\resizebox{!}{.0285\paperheight}{\begin{tabular}{ccccc}

\hline
\multirow{2}{*}{Bands} & \multirow{2}{1.3cm}{Faster RCNN}  & \multirow{1}{2.5cm}{MLMLT-CNN (Early - addition)} & \multirow{1}{2.3cm}{MLMLT-CNN (Early - concat.)} & \multirow{1}{2.3cm}{MLMLT-CNN (Late - concat.)}  \\ \\

\hline
{T1Gd}  & {0.73}  & {0.74}  & {0.83}  & \textbf{{0.89}}  \\
{T1}    & {0.54}  & {0.78}  & {0.89}  & \textbf{{0.91}}  \\
{T2}    & {0.56}  & {0.76}  & {0.86}  & \textbf{{0.89}}  \\
{Flair} & {0.48}  & {0.75}  & {0.86}  & \textbf{{0.91}}  \\

\hline
\end{tabular}
} 
\end{table}

When comparing the detection results per image band, we notice that 304~{\AA} images are repeatedly amongst the most difficult to analyse in UAD, having the lowest F1-scores in all tests.
On the other hand, 171~{\AA} shows the highest results of all UAD bands, followed by 284~{\AA} and 195~{\AA}, respectively.
This may be explained by ARs having a denser or less ambiguous appearance in 171~{\AA}, 195~{\AA}, and 284~{\AA} image bands than in 304~{\AA} since they are higher in the corona.
A similar observation can be made in the LAD dataset when comparing the Magnetogram results to PM/SH 3934~{\AA}, where Magnetograms observe a lower altitude than PM/SH 3934~{\AA}.
This demonstrates that the these bands are not equal in how difficult they may be analysed, even though they were acquired at the same time with same size and resolution.
These observations suggest that detecting ARs using information provided by a single band may be an under-constrained problem.

\subsubsection{Joint detection on multiple image bands}
\label{JointDet}

We now present the results of our framework when detecting ARs over the UAD bands jointly. We experiment with different types of feature fusion and different combinations of bands.
We compare against the state-of-the-art AR detector HFC's SPOCA \cite{method:spoca}.
We further compare against a sequential fine-tuning method derived from \cite{paper:mul-ch-coronal-hole-det} through adapting the first stage of their approach to Faster RCNN by sequentially fine tuning it over the neighbouring image bands. 
We evaluate this approach on UAD.
Moreover, we compare against Faster RCNN on single bands to demonstrate the benefit of jointly processing the image bands, taking into account their inter-dependencies for more robust individual detections.

\begin{figure}[t]
\centering
\begin{tabular}{c}

    \includegraphics[width=0.95\linewidth]{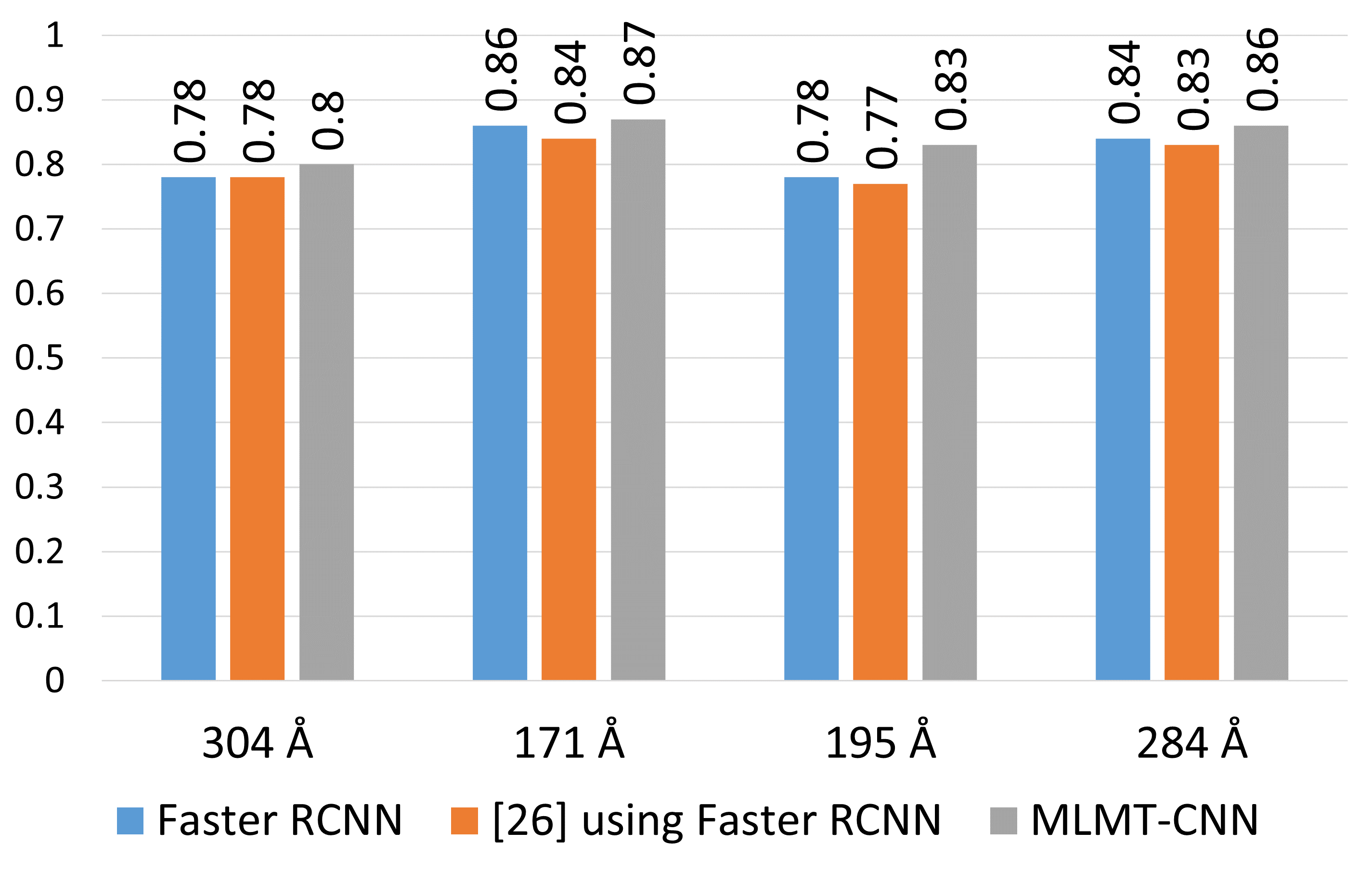} \\

    \includegraphics[width=0.95\linewidth]{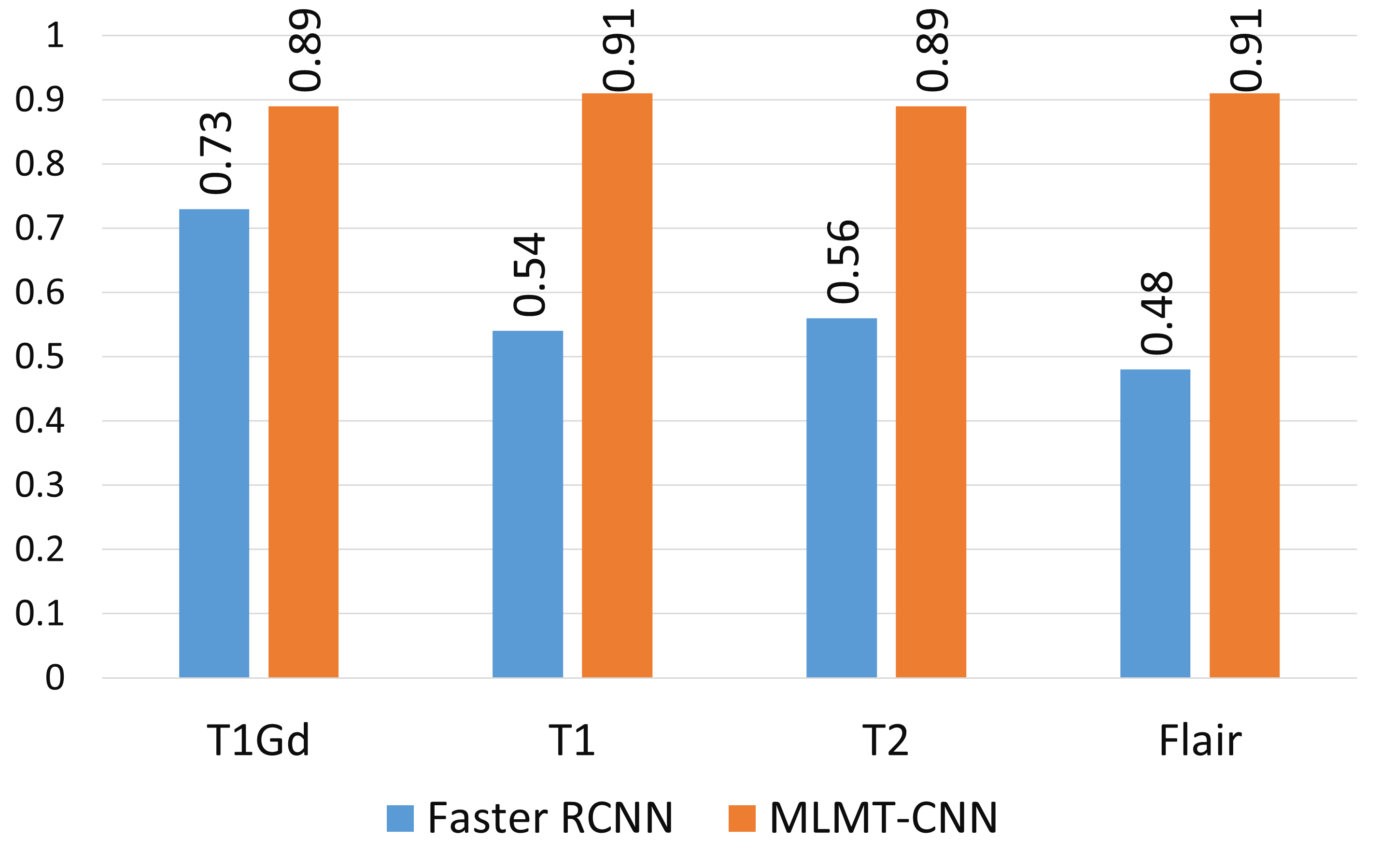} \\
    
    \end{tabular}

    \caption{{Comparison of the detection results over UAD (top) and BraTS-prime (bottom) datasets. Each group of bars represents an imaging modality. Different colors represent different methods.}}

\label{fig:detection_barchart_comparision}
\end{figure}

\begin{figure}[!ht]
	\centering
		\includegraphics[width=1.0 \linewidth]{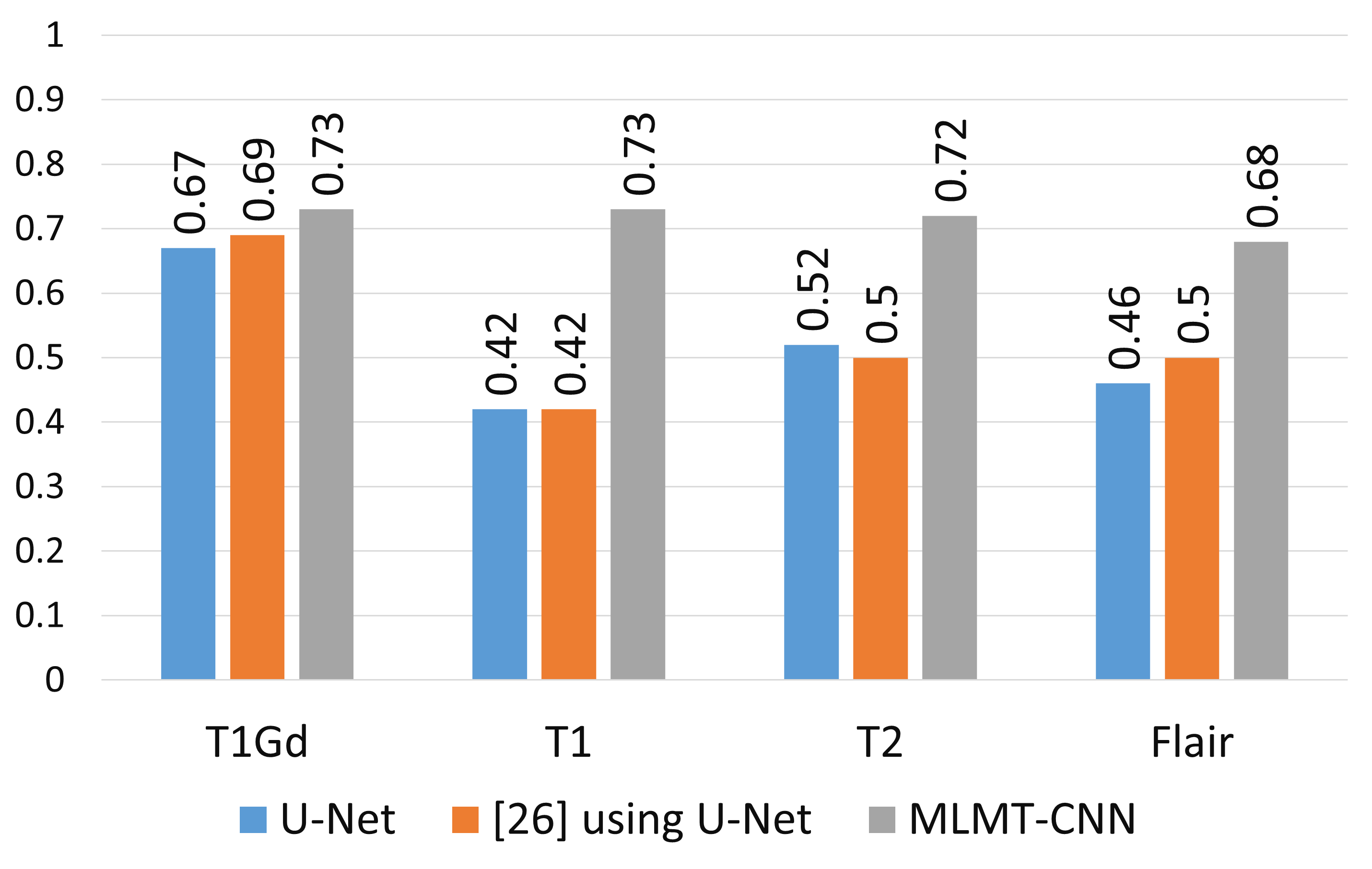}

    	\caption{{Comparison of the segmentation results over BraTS-prime dataset. Each group of bars represents an imaging modality. Different colors represent different methods.}}
	\label{fig:segmentation_barchart_comparision}
\end{figure}

\begin{table}
\centering
\caption{Performance of single image segmentation over BraTS-prime. For each class, the highest scores are highlighted in bold.}
\label{tab:eval_seg_single_brats}
\resizebox{!}{.068\paperheight}{
\begin{tabular}{ccccccc}
\hline
\multirow{2}{*}{Architecture} & \multirow{2}{*}{Supervision} & \multirow{2}{*}{Bands} & \multicolumn{3}{c}{IoU score per class} & Mean   \\
\hhline{~~~---}
& & & NCR/NET  & ED  & ET  &    IoU      \\

\hline
\multirow{4}{*}{FCN8}  & \multirow{4}{1.2cm}{\centering{Fully supervised}} & T1Gd &    0.54      &     0.43    &  0.70     & 0.56  \\
\hhline{~~-----}
 & &  T1 & 0.08 & 0.33 & 0.0 & 0.14 \\
\hhline{~~-----}
 & &  T2 & 0.49 & 0.48 & 0.23 & 0.40 \\
\hhline{~~-----}
 & &  Flair& 0.43 & 0.51 & 0.19 & 0.38 \\

\hline
\multirow{4}{*}{U-Net}  & \multirow{4}{1.2cm}{\centering{Fully supervised}} &  T1Gd &    \textbf{0.69}     &     0.52    &   \textbf{0.80}   & \textbf{0.67}  \\
\hhline{~~-----}
 & &  T1 & 0.56 & 0.50 & 0.19  & \textbf{0.42} \\
\hhline{~~-----}
 & & T2 & 0.63 & 0.56 & 0.36 & \textbf{0.52} \\
\hhline{~~-----}
 & & Flair& 0.50 & \textbf{0.59} & 0.29 & \textbf{0.46} \\

\hline
\multirow{4}{*}{U-Net}  & \multirow{4}{1.2cm}{\centering{Weakly supervised}}  &  T1Gd &      0.66    &     0.33    &  0.53     & 0.51  \\
\hhline{~~-----}
 & &  T1 & 0.58 & 0.39 & 0.0 & 0.32 \\
\hhline{~~-----}
 & &  T2 & 0.58 & 0.43 & 0.1 & 0.37 \\
\hhline{~~-----}
 & &  Flair& 0.44 & 0.49 & 0.0 & 0.31 \\

\hline
\end{tabular}
} 
\end{table}

\begin{table}
\centering
\caption{Segmentation performance of MLMT-CNN (U-Net) with full supervision over BraTS-prime for different numbers of modalities and feature fusions. For each class, the highest scores are highlighted in bold.}
\label{tab:eval_seg_multi_brats}
\resizebox{!}{.165\paperheight}{
\begin{tabular}{cccccccc}
\hline
Archi- & \multirow{2}{*}{Fusion} & Slice & \multirow{2}{*}{Bands} & \multicolumn{3}{c}{IoU score per class} & Mean  \\
\hhline{~~~~---}
 tecture &     &  gap  &      & NCR/NET  & ED  & ET  &     IoU     \\

\hline
\multirow{29}{1.2cm}{MLMT-CNN (U-Net)}     &     \multirow{9}{0.6cm}{Early - concat.} & \multirow{9}{*}{1}   & T1 &    0.60      &     0.56    &  0.41 & 0.52  \\
& & & T2 & 0.59 & 0.59 & 0.39 & 0.52 \\

\hhline{~~~-----}
     &       &     & T1 &    0.62      &     0.59    &  0.35 & 0.52  \\
& & & T2 & 0.63 & 0.61 & 0.36 & 0.53 \\
& & & Flair & 0.63 & 0.63 & 0.39 & 0.55 \\

\hhline{~~~-----}
    &      &   & T1Gd &    \textbf{0.75}      &     \textbf{0.66}    &  0.78 & \textbf{0.73}  \\
& & & T1 & \textbf{0.75} & \textbf{0.69} & \textbf{0.76} & \textbf{0.73} \\
& & & T2 & \textbf{0.74} & \textbf{0.71} & \textbf{0.70} & \textbf{0.72} \\
& & & Flair& \textbf{0.73} & \textbf{0.68} & 0.62 & \textbf{0.68} \\

\hhline{~-------}
    &     \multirow{4}{0.6cm}{Early - addition} & \multirow{4}{*}{1}   & T1Gd &    0.74      &     0.64    &  0.78 & 0.72  \\
& & & T1 & 0.73 & 0.67 & 0.74 & 0.71 \\
& & & T2 & 0.69 & 0.66 & 0.65 & 0.67 \\
& & & Flair& 0.68 & 0.67 & 0.61 & 0.65 \\

\hhline{~-------}
    &     \multirow{4}{0.6cm}{Late - concat.} & \multirow{4}{*}{1}   & T1Gd &    0.71      &     0.63    &  0.81 & 0.72  \\
& & & T1 & 0.73 & 0.65 & 0.74 & 0.71 \\
& & & T2 & 0.71 & 0.68 & \textbf{0.70} & 0.70 \\
& & & Flair& 0.69 & 0.67 & \textbf{0.64} & 0.67 \\

\hhline{~-------}
    &     \multirow{4}{0.6cm}{Late - addition} & \multirow{4}{*}{1}   & T1Gd &    0.70      &     0.60    &  0.81 & 0.70  \\
& & & T1 & 0.71 & 0.63 & 0.73 & 0.69 \\
& & & T2 & 0.67 & 0.66 & 0.67 & 0.67 \\
& & & Flair& 0.68 & 0.66 & 0.60 & 0.65 \\

\hhline{~-------}
    &     \multirow{8}{0.6cm}{Early - concat.} & \multirow{4}{*}{2}   & T1Gd &    0.68      &     0.55    &  0.76 & 0.66  \\
& & & T1 & 0.67 & 0.61 & 0.66 & 0.65 \\
& & & T2 & 0.64 & 0.65 & 0.55 & 0.61 \\
& & & Flair& 0.57 & 0.62 & 0.46 & 0.55 \\

\hhline{~~------}
     &       & \multirow{4}{*}{3}   & T1Gd &    0.63      &     0.51    &  0.73 & 0.62  \\
& & & T1 & 0.63 & 0.60 & 0.62 & 0.62 \\
& & & T2 & 0.57 & 0.64 & 0.43 & 0.55 \\
& & & Flair& 0.59 & 0.61 & 0.41 & 0.54 \\

\hline
\multirow{4}{1.2cm}{\cite{paper:mul-ch-coronal-hole-det} using U-Net }     &     \multirow{4}{0.6cm}{Sequential fine-tuning} & \multirow{4}{*}{1}   & T1Gd &    0.71      &     0.55    &  \textbf{0.82} & 0.69  \\
& & & T1 & 0.56 & 0.50 & 0.19 & 0.42 \\
& & & T2 & 0.65 & 0.58 & 0.26 & 0.50 \\
& & & Flair& 0.57 & 0.61 & 0.33 & 0.50 \\

\hline
\end{tabular}
} 
\end{table}

\begin{table}
\centering
\caption{Evaluation of weakly supervised MLMT-CNN (U-Net) on BraTS-prime.
For each class, the highest scores are highlighted in bold.}
\label{tab:eval_seg_multi_weak_brats}
\resizebox{!}{.075\paperheight}{
\begin{tabular}{cccccc}
\hline
\multirow{2}{1.2cm}{\# train. stages}  & \multirow{2}{*}{Bands} & \multicolumn{3}{c}{IoU score per class} & Mean \\
\hhline{~~---}
 &          & NCR/NET  & ED  & ET  & IoU \\

\hline
\multirow{4}{*}{1} & T1Gd &    0.67      &     0.40    &  0.38 & 0.48  \\
& T1 & 0.66 & 0.41 & \textbf{0.40} & 0.49 \\
& T2 & 0.62 & 0.45 & 0.39 & 0.49 \\
& Flair& 0.64 & 0.46 & 0.38 & 0.49 \\

\hline
\multirow{4}{*}{2} & T1Gd &    \textbf{0.69}      &     0.43    &  \textbf{0.40} & \textbf{0.51}  \\
 & T1 & \textbf{0.69} & 0.41 & \textbf{0.40} & \textbf{0.50} \\
 & T2 & 0.66 & 0.45 & 0.38 & \textbf{0.50} \\
 & Flair& 0.67 & \textbf{0.47} & 0.38 & \textbf{0.51} \\

\hline
\multirow{4}{*}{3} & T1Gd &   0.67 & 0.40 
& 0.37 & 0.48  \\
 & T1 & 0.67 & 0.40 & 0.37 & 0.48 \\
 & T2 & 0.64 & 0.42 & 0.36 & 0.47 \\
 & Flair& 0.64 & 0.45 & 0.34 & 0.48 \\

\hline
\end{tabular}
} 
\end{table}

\begin{table}
\centering
\caption{
Comparison of full and weak supervision for MLMT-CNN (U-Net) over weak-Cloud-38. For each band, the highest scores of the weakly-supervised models are highlighted in bold.}

\label{tab:eval_seg_multi_cloud_38}
\resizebox{!}{0.036\paperheight}{
\begin{tabular}{ccccccc}
\hline
Super- & \multirow{2}{1.2cm}{\# train. stages} & \multicolumn{4}{c}{IoU score per band} & Mean \\
\hhline{~~----~}
vision &   & Red & Green & Blue & NIR & IoU   \\

\hline
Fully    & NA  &    0.95      &     0.95    &  0.95       & 0.95 & 0.95  \\

\hline
 \multirow{3}{*}{\centering{Weakly}}     &  1 &  0.78      &     \textbf{0.80}    &  \textbf{0.83}       & \textbf{0.83} & \textbf{0.81}  \\

    &  2   &    \textbf{0.79}      &     \textbf{0.80}    &  \textbf{0.83}       & \textbf{0.83}    & \textbf{0.81}  \\

 &   3   &    0.78      &     \textbf{0.81}    &  0.82       & \textbf{0.83}    & \textbf{0.81}  \\

\hline
\end{tabular}
} 
\end{table}

In our first experiment, we compare early fusion (pixel level  concatenation) against late fusion (feature level concatenation or addition), on the LAD dataset. Overall, the three approaches show an enhanced performance in contrast to single band based detection.
However, we find that late fusion with concatenation shows higher performance than early fusion, having 0.90 F1-score versus 0.88 for magnetograms, while both scored 0.89 over 3934~{\AA}. We further test late fusion using element wise addition and observe a decrease of 1\% and 3\% in the F1-score over 3934~{\AA} and Magnetogram, respectively.
Late fusion is thus adopted for all following experiments. 

We also evaluate the benefit of our MOO strategy using our 2-band based architecture on the UAD dataset. As seen in Table \ref{table:eval_SRCNN_Raw_images}, this approach generally improves the F1-scores in most bands comparing to the non-MOO architectures. This behaviour may indicate that the two feature extraction stages were indeed more effectively optimised for their different tasks at different epochs. Thus we use this MOO approach for all other experiments.

On the UAD dataset, with various combinations of 2 bands, we notice a general improvement over single band detections. In addition, the performance varies in correspondence to the bands being used. Combining bands that are difficult to analyse (304~{\AA} or 195~{\AA} that have lowest F1-scores in the single band analyses) with easier bands (171~{\AA} and 284~{\AA}) unsurprisingly enhances their respective performance. More interestingly, combining the difficult 304~{\AA} and 195~{\AA} bands together also improve on their individual performance. Similarly, when combining bands that are easier to analyse (171~{\AA} and 284~{\AA}), performances are also improved over their individual analyses. Following these settings, our 2-band based approach was able to record higher or similar F1-scores in contrast to the best performing single-band detector. This supports our hypothesis that joint detection may provide an increased robustness through learning the inter-dependencies between the image bands. Moreover, the most dramatic improvement in F1-scores across both LAD and UAD datasets is for the 3934~{\AA} images when magnetograms are added to the analysis. This is in line with the current understanding of AR having strong magnetic signatures.

Generally, in the UAD dataset, we find that using a combination of 2 bands produces the best F1 scores in comparison to using 3 or 4 bands in the analysis, see Table \ref{table:eval_SRCNN_Raw_images}.
This may be caused by the fact that optimising the network for multiple tasks (2, 3, or 4 detection tasks) simultaneously increases the complexity of the problem. While the network successfully learned to produce better detections in the case of 2 bands, it was difficult to find a generalised yet optimal model for 3 or 4 bands at the same time. Thus, for 4 bands, the model obtains the best precision but at the expense of a poor recall.

\begin{figure}[!ht]
	\centering
		\includegraphics[width=0.9 \linewidth]{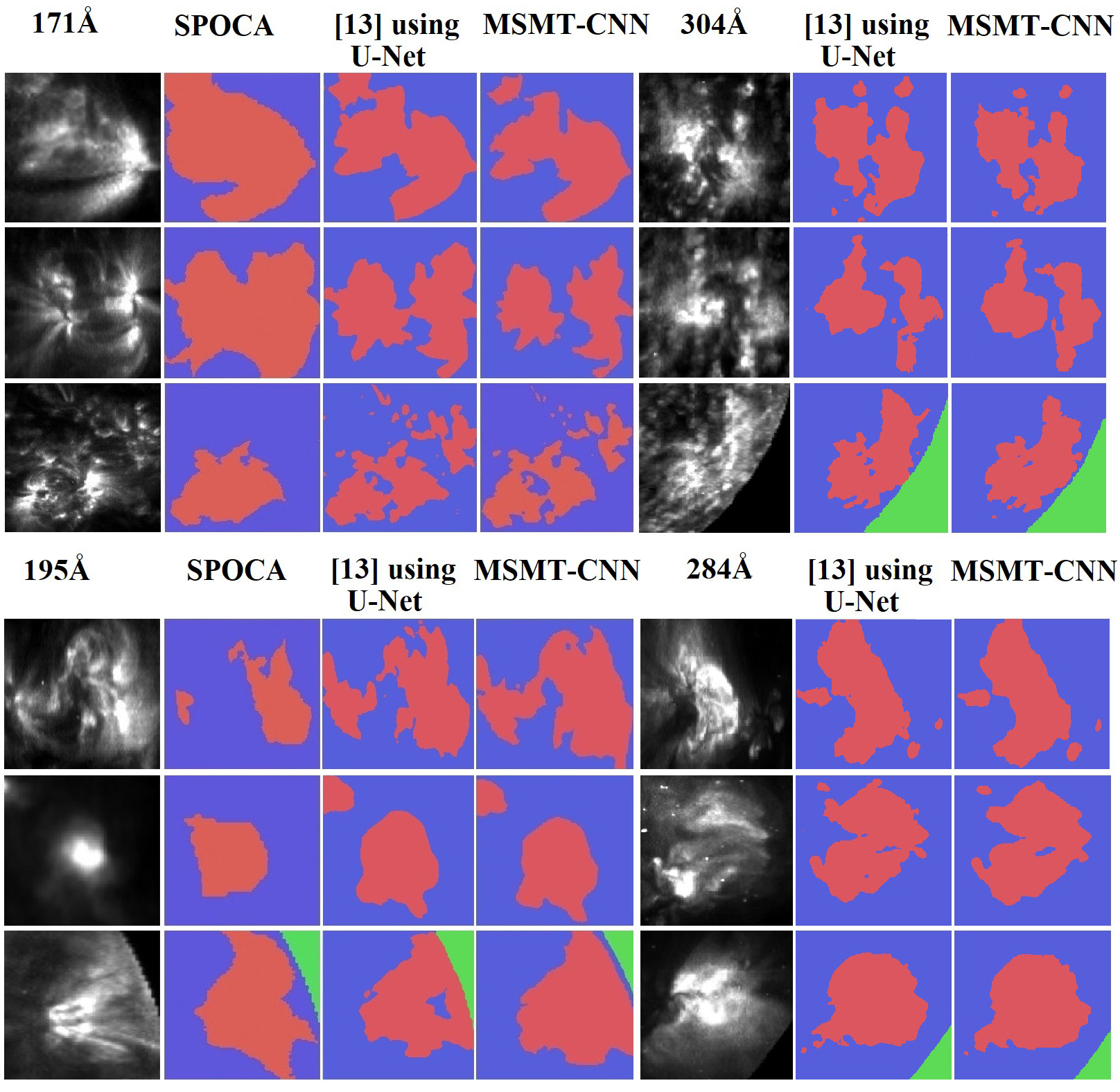}
	\caption{AR segmentation comparison between our presented method, SPOCA, and sequentially fine-tuned DNNs similar to \cite{paper:mul-ch-coronal-hole-det}, over the SPOCA subset. Red is AR, blue denotes the quite Sun background, and green is outside of the solar disk.}
	\label{fig:SPOCA_vs_SRCNN_vs_MulChCorHolDet_segmentation}
\end{figure}

On the SPOCA subset, over the bands 171~{\AA} and 195~{\AA} for which it is designed, the SPOCA method obtains the poorest performance of all multi-band and single-band experiments.
It is worth noting that this method relies on manually tuned parameters according to the developers' own definition and interpretation of AR boundaries, which may differ from the ones we used when annotating the dataset. While supervised DL-based methods could integrate this definition during training, SPOCA could not perform such adaptation. This may have had a negative impact on its scores. Furthermore, visual inspection shows a poor performance for SPOCA on low solar activity images, see Fig.~\ref{fig:UAD_activity_levels}.
This may be due to the use of clustering in SPOCA, since in low activity periods the number of AR pixels (if any) is significantly smaller than solar background pixels, which makes it difficult to identify clusters.

Moreover, the sequential fine tuning approach similar to \cite{paper:mul-ch-coronal-hole-det} shows a close performance to single band detection using Faster RCNN with an identical precision, recall and F1-score over the band 304~{\AA} and a slight decrease over the other 3 bands, See Table \ref{table:eval_SRCNN_Raw_images} and Fig. \ref{fig:detection_barchart_comparision}. This may be due to the fact that its transfer learning does not incorporate the bands' inter-dependencies when analysing the different bands. Moreover, the method was designed in \cite{paper:mul-ch-coronal-hole-det} to produce a single prediction for the different bands, this differs from our usage 
where we predict a different set of detections per band.

We further evaluate our detection approach with different fusions, over the 4 bands of BraTS-prime dataset, and compare it against single band based detection. 
All fusion strategies significantly outperform single band detectors, with late concatenation fusion being the highest, showing an average F1-score increase of ~39\% across all modalities. See Table \ref{table:brats_det_results_short} and Fig. \ref{fig:detection_barchart_comparision}. This confirms our hypothesis that exploiting inter-dependencies between the image bands by the joint analysis may provide a superior performance in contrast to single band based detection.

\subsection{Segmentation stage}
\label{sub:Segemtation-stage-results}

Our AR segmentation results were all qualitatively assessed and validated by a solar physics expert. We also visually compare the results against SPOCA and a sequentially fine-tuned U-Net model (similar to the first stage of \cite{paper:mul-ch-coronal-hole-det}).

Additionally, 
to quantitatively demonstrate the benefit of the joint analysis,
and due to the lack of manual AR pixel-wise ground-truth, we evaluate our approach using 
the BraTS-prime synthetic dataset.
Weak-Cloud-38 may not be used for this purpose because of its different bands capturing the same scene, rather than different layers of a 3D object.
It is worth noting that we do not aim to achieve state-of-the-art performance in tumour segmentation, but rather to confirm the benefit of the joint analysis in scenarios similar to our solar case, where different modalities show different cuts of a 3D object.
Since ground-truth is available for this dataset, we follow the classical fully-supervised training procedure.
Furthermore, we use Weak-BraTS-prime and Weak-Cloud-38 to evaluate our iterative training strategy from weak labels against full supervision.

Its worth noting that the segmentation subnetworks adopt the same layers configuration of their correspondent blocks in U-Net \cite{paper:UNet}.

\subsubsection{Independent segmentation on single image band}
\label{independentSeg}

We first compare segmentation results produced by U-Net and FCN8 over the AR
and BraTS-prime (Table \ref{tab:eval_seg_single_brats}) individual image bands, analysed independently, to evaluate different DL-based segmentation networks. These results also serve as baseline to assess our joint analysis based approach in Section \ref{JointSeg}.

We notice that U-Net produces higher IoU values over all bands for BraTS-prime,
as well as smoother AR boundaries, compared to FCN8.
This is expected since U-Net utilises skip connections to help retrieving fine details in the mask reconstruction process. Therefore, we use the building blocks of U-Net in our joint segmentation framework.

When comparing the results of U-Net over different modalities, we notice that the T1-Gd modality gets the highest IoU score for the ET class. A similar trend can be seen when comparing the results of the NCR/NET class over different modalities. On the other hand, we find that Flair gets the highest IoU for the ED class comparing to the other modalities. This contrast in the IoU scores is in line with the understanding that different modalities provide different information.

\subsubsection{Joint segmentation on multiple image bands}
\label{JointSeg}

Similar to our detection experiment, we assess our framework using different combinations of image bands and different types of feature fusion to evaluate their influence on the segmentation performance.

\paragraph{Quantitative results}
\label{Quantitative_results_Seg}

First, we present our BraTS-prime segmentation on combined bands using our joint analysis approach (Table \ref{tab:eval_seg_multi_brats}). 
We note that all combinations improve on the single-band results, 
with the 
best improvement coming from combining all four modalities. 
All following BraTS-prime experiments use a four-band architecture.

We compared four fusion strategies, namely fusing features after one block of convolution only (early) and at the end of convolutions (late), using addition and concatenation. We find that early fusion with concatenation shows higher results.
This differs from our observation in the AR detection experiment, hence confirming that the fusion strategy needs to be adapted to the analysis scenario.
Accordingly, we continue using early fusion with concatenation for all BraTS-prime segmentation experiments.

As expected, there is a negative correlation between the IoU score and the width of slice gap, where the overall increase in the IoU was the highest for smaller gaps and higher levels of spatial correlation (gap of 1 pixel). This observation, together with the improved results from combining bands, suggest that jointly analysing related multi-modal images in scenarios similar to our solar case may indeed aid the network in learning the inter-dependencies between the different modalities.

We compare against sequentially fine-tuned U-Net models similar to the first stage of \cite{paper:mul-ch-coronal-hole-det} in Table \ref{tab:eval_seg_multi_brats} and Fig. \ref{fig:segmentation_barchart_comparision}. They achieved comparable IoU scores to those produced by U-Net on single bands. Hence, they do not benefit from the combination of modalities as our framework does.

Additionally, as a mean to assess our iterative training steps,
we use weak-BraTS-prime and weak-Cloud-38 to evaluate this strategy
against manual annotations, and compare it to the classical training approach.

When evaluating the recursively trained model using weak-BraTS-prime dataset against the fully supervised model on BraTS-prime manual annotations,
we notice an increase in the IoU scores after one step of recursion (i.e. 2 stages of training, first using the weak labels, then using the previous predictions as labels), achieving 71\% of the fully supervised performance (Table \ref{tab:eval_seg_multi_weak_brats}).
Moreover, this iterative training process achieves 85\% of the fully supervised approach over the Weak-Cloud-38 dataset, with the best performance also being after one round of recursion, with an increase of 1\% over the Red band (Table \ref{tab:eval_seg_multi_cloud_38}).
These observations indicate that our recursive training strategy is beneficial in cases where manual annotations are not available, such as solar ARs.

In contrast to the single band based segmentation of weak-BraTS-prime (last 4 rows of Table \ref{tab:eval_seg_single_brats}), we also note that performance still benefits from the joint analysis even when trained -- classically or recursively -- with weak labels (Table \ref{tab:eval_seg_multi_weak_brats}).

\paragraph{Qualitative results}
\label{Qualitative_results_Seg}

Lastly, we compare visually our segmentation results on the SPOCA subset, using our proposed architecture, against SPOCA and sequentially fine-tuned DNNs similar to \cite{paper:mul-ch-coronal-hole-det} (without their final stage of fusing the CNNs' individual predictions) (Fig.~\ref{fig:SPOCA_vs_SRCNN_vs_MulChCorHolDet_segmentation}).
The results show that our framework generally finds more detailed AR shapes than SPOCA, while at the same time being more robust to fainter regions of ARs.

Additionally, we compare our AR segmentation results to SPOCA by finding the IoU between the predictions produced by the two approaches over the SPOCA subset.
This may be used to indicate the agreement between the two methods.
We find that both 171~{\AA} and 195~{\AA} achieve
a higher agreement 
of 44\% and 46\%, respectively, in contrast to 304~{\AA} and 284~{\AA} scoring 33\% and 41\%, respectively.
This is expected since SPOCA was designed to segment ARs in 171~{\AA} and 195~{\AA}.
Overall, the similarity between our predictions and SPOCA's is relatively low. However, as discussed in Section \ref{JointDet}, SPOCA was manually tuned by the developers according to their own interpretation of AR boundaries which may be different from our interpretation when annotating the dataset. Hence, care must be taken when interpreting the results.

Comparison against sequentially fine-tuned CNNs in the spirit of \cite{paper:mul-ch-coronal-hole-det} is fairer, since the DNNs were trained on our data. Segmentation of the sequentially fine-tuned CNNs appears to be of similar quality to ours, although shapes of an AR between neighbouring bands evolve more smoothly with our method.
This is an advantage of accounting for the 3D geometry of ARs in performing the 2D segmentation.

\section{Conclusion} 
\label{sec:concl}
We presented a multi-layer and multi-tasking framework to tackle the 3D solar AR detection and segmentation problem from multi-spectral images that observe different layers of the 3D solar atmosphere. MLMT-CNN analyses multiple bands jointly to produce consistent localisation. It is a flexible framework that may use different CNN backbones, and may be generalised to any number and modalities of images.
We find that by fusing information from different image bands at different feature levels, CNNs were able to localise objects more robustly and more consistently across layers. Additionally, our study suggests that different imaging scenarios may require different types of feature fusion strategies. We also show that the number of bands used in the analysis might affect the performance and must be optimised to each case.
Furthermore, we demonstrate that CNNs may show a satisfactory localisation performance when iteratively trained from weak annotations.
MLMT-CNN showed competitive results against both baseline and state-of-the-art detection and segmentation methods.
Future research could investigate the information importance of different image bands and its influence on task learning in both multi-spectral and multi-layer scenarios.

\bibliographystyle{unsrt} 
\bibliography{refs}

\end{document}